\documentclass[10pt,twocolumn]{IEEEtran}

\pdfoutput=1

\ifCLASSINFOpdf
  \usepackage[pdftex]{graphicx}
\else
  \usepackage[dvips]{graphicx}
\fi
\usepackage{algorithmic}
\usepackage{amsmath}
\usepackage{amssymb}
\usepackage{multirow}
\usepackage{rotating}
\usepackage{booktabs}
\usepackage{multirow}
\usepackage{gensymb}
\usepackage{algorithm}

\begin{document}
%
\title{Detect2Rank : \\ Combining Object Detectors Using \\ Learning to Rank}
%
%
%

\author{Sezer~Karaoglu,
        Yang~Liu
        and Theo~Gevers,~\IEEEmembership{Member,~IEEE,}
\thanks{S. Karaoglu is with the Intelligent Systems Lab, Amsterdam, University of Amsterdam, 1098 XH Amsterdam, The Netherlands (e-mail: s.karaoglu@uva.nl). Y. Liu is with the Intelligent Systems Lab, Amsterdam, University of Amsterdam, 1098 XH Amsterdam, The Netherlands. T. Gevers is with the Intelligent Systems Lab, Amsterdam, University of Amsterdam, 1098 XH Amsterdam, The Netherlands, and also with the Computer
Vision Center, Universitat Aut\`{o}noma de Barcelona, 08193 Barcelona, Spain
(e-mail: th.gevers@uva.nl)}}

\maketitle

\begin{abstract}

Object detection is an important research area in the field of computer vision. Many detection algorithms have been proposed. However, each object detector relies on specific assumptions of the object appearance and imaging conditions. As a consequence, no algorithm can be considered as universal. With the large variety of object detectors, the subsequent question is how to select and combine them.

In this paper, we propose a framework to learn how to combine object detectors. The proposed method uses (single) detectors like DPM, CN and EES, and exploits their correlation by high level contextual features to yield a combined detection list.

Experiments on the PASCAL VOC07 and VOC10 datasets show that the proposed method significantly outperforms single object detectors, DPM (8.4\%), CN (6.8\%) and EES (17.0\%) on VOC07 and DPM (6.5\%), CN (5.5\%) and EES (16.2\%) on VOC10.

\end{abstract}

\begin{IEEEkeywords}
Object Detection, Fusion, Learning to rank
\end{IEEEkeywords}

%
\IEEEpeerreviewmaketitle

\section{Introduction}

\IEEEPARstart{O}{bject} detection is an active research area in the field of computer vision. Many detection algorithms have been proposed~\cite{Overfeat,DPM,SSER,MKL,CNHOG,EES,jacob,RCNN}. Although these detection algorithms are successful for many detection tasks, they may be less accurate for some specific cases.

To gain more insight on the differences amongst detectors, Hoiem et al.~\cite{Diagnose} provide an extensive analysis on object detectors and their properties~\cite{Diagnose}. Their findings are that detectors perform well for standard object appearances and for common imaging conditions. Obviously, different design properties of the detectors (e.g. search strategy, features, and model presentation) influence the robustness of the methods to varying imaging conditions (e.g. occlusion, clutter, unusual views, and object size). For instance, detectors based on the sliding-window approach~\cite{DPM} using pre-defined window sizes and aspect ratios are good at finding likely object positions (rough object positions). However, they are less suited to detect deformable objects precisely. Hoeim et al.~\cite{Diagnose} show that these type of detectors typically suffer from poor localization errors. On the other hand, a flexible sliding-window allows detecting deformable objects. The large number of candidate regions to be considered for detection limits the use of strong classifiers. Therefore, selective search~\cite{SSER} is integrated as a pre-processing step in current state-of-the-art techniques~\cite{RCNN} to reduce the computational complexity of sliding-window based approaches to generate a reduced set of candidate regions. However, Hosang et al.~\cite{Hosang} show that selective search generates candidate regions which are sensitive to changes in scale, illumination and geometrical transformations. This is because selective search is based on segmentation derived from superpixels which are unstable for small image deformations.

Besides the method to generate proper candidate regions for detection, the choice of features influences the robustness and discriminative power of the detectors. HOG-based templates are able to preserve shape information~\cite{DPM,EES} of the objects but are less suited to differentiate between visually similar categories such as cats and dogs. This limitation is addressed using color information in~\cite{CNHOG}, following successful results of using color information in object recognition~\cite{Color}. HOG-based object detection using color~\cite{CNHOG} is suited for object classes in which the intra class color variation is low (e.g. potted plant and tv-monitor). However, the use of color negatively affects the detection accuracy for object classes in which the intra class color variation is large (e.g. bottles and buses).

Finally, the chosen model and classifier drastically influences the performance of the detectors. In general, object detectors represent all positive samples of a given category as a whole~\cite{DPM,CNHOG}. However, Malisiewicz and Efros~\cite{Model} show that standard categories (e.g. train, car and bus) do not form coherent visual categories. Accordingly these methods are too generic. To address this issue Malisiewicz et al.~\cite{EES} propose to train a separate linear SVM classifier for each positive sample in the training set. Gu et al.~\cite{Multi} show that using only one positive sample for training significantly reduces the generalization capacity. Hence, the detection performance of~\cite{EES} is deteriorated for uncommon object views.

\begin{figure*}[t]
\begin{center}
   \includegraphics[width=.9\linewidth]{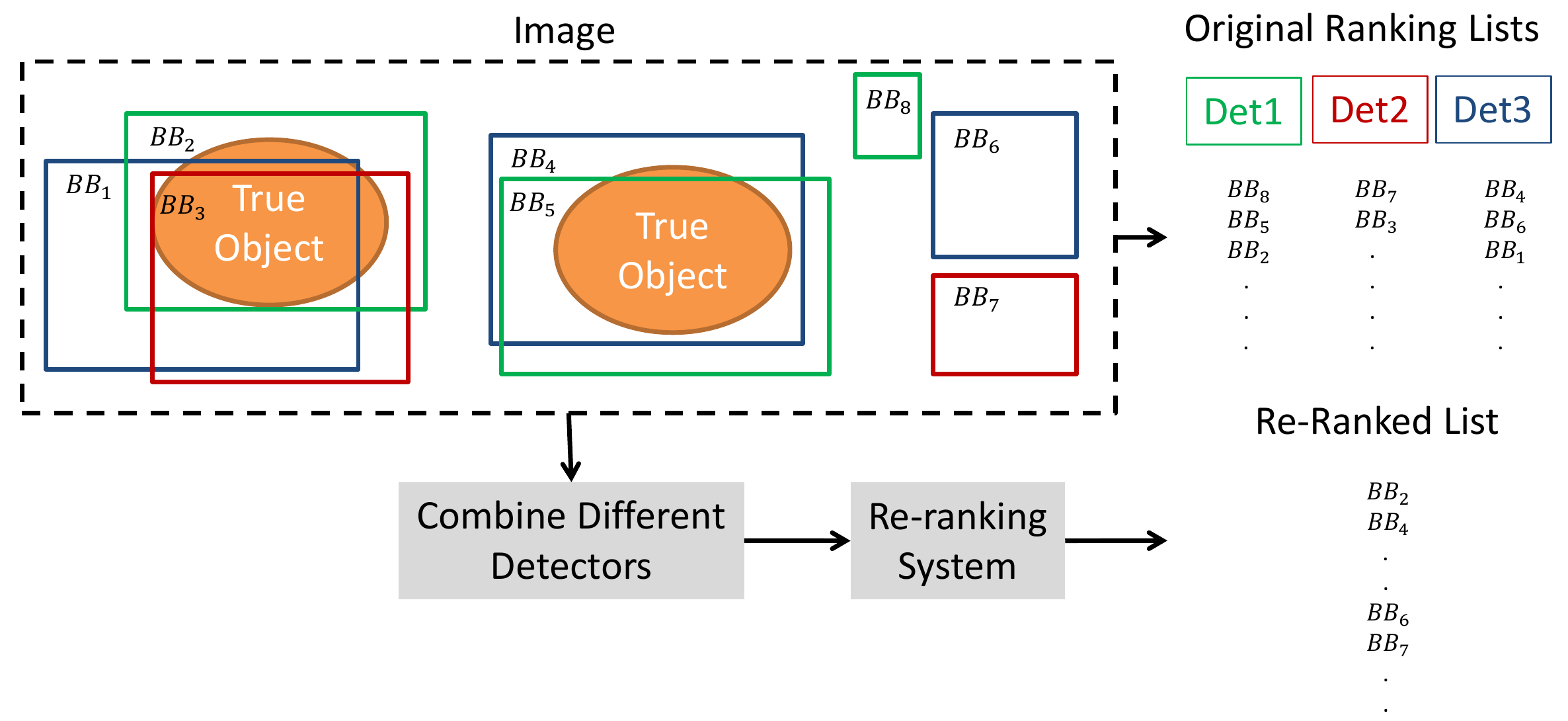}
\end{center}
   \caption{Flow of the proposed method (Best viewed in color). Initial detections from different detectors namely, Det1(green), Det2(red) and Det3(Blue) are combined by a learning to rank algorithm. False detections of the individual detectors are learned by detector-detector relations and obtain less confidence when combined, whereas consistency in detectors $BB_1$, $BB_2$ and $BB_3$ are rewarded by the re-ranking system.}
\label{fig:flow}
\end{figure*}


As a consequence, no detection algorithm can be considered as universal. With the large variety of available methods, the question is how to combine these object detectors preserving their strengths while reducing their limitations and assumptions. In this paper, we consider a rank learning approach to combine object detection methods. The proposed framework combines detections (detector outputs which consist of a classifier score and bounding box locations) of different state-of-the-art object detectors including DPM~\cite{DPM}, CN~\cite{CNHOG} and EES~\cite{EES}. Furthermore, the method extracts high-level context features such as detector-detector consistency, detector-class preference, object-saliency of a detection, and object-object relations. These features are used in a learning to rank framework to yield a combined detection list. The flow of the proposed method is summarized in Fig~\ref{fig:flow}.

The proposed approach offers the following advantages over single object detectors:

\begin{itemize}
  \item Missed detections (false negatives) of single detectors are compensated by combining detections of different detectors.
  \item Detections are re-ranked by using information gathered by other detectors. True detections (true positives) of each detector are rewarded and false detections (false positives) of each detector are penalized within the learning to rank framework.
  \item The combined list maintains the strengths of the detectors. Therefore, it is more robust than each single detector for varying imaging conditions.
\end{itemize}


To the best of our knowledge, we are the first to propose using re-ranking approaches to combine object detectors. Experiments on the PASCAL VOC07 and VOC10 datasets show that the proposed method significantly outperforms single detectors. \label{sec:intro}

\section{Related Work}\label{relatedWork}


\subsection{Object Detection}
In general, papers on object detection aim at designing a single detector, descriptor or classifier~\cite{Overfeat,DPM,MKL,RCNN,jacob,SS,Deep}. Felzenszwalb et al.~\cite{DPM} propose a part-based object detection method using HOG features and a latent SVM. This algorithm outperforms the state-of-the-art methods for standard object appearances. The use of template-based models limits the capability to detect deformable objects~\cite{Diagnose}. Moreover, template-based models (using HOG features) are designed to accommodate for shape information and are less suited to differentiate visually similar categories (e.g cats and dogs). In contrast to part-based detection methods, Vedaldi et al.~\cite{MKL} propose to use a bag-of-words model for object detection. Multiple features are used within a multiple kernel learning framework which is able to distinguish between visually similar object categories. However, Hoiem et al.~\cite{Diagnose} show that this approach is sensitive to object size due to the bag-of-words model. Khan et al.~\cite{CNHOG} propose to use additional color information for object detection. The color information contains expressive power for object classes in which the intra class color variations are low (e.g. potted-plants and sheeps). However, color may have a negative influence on the detection of classes in which the intra class color variations are high (e.g. bottles and buses)~\cite{CNHOG}.

Malisiewicz et al.~\cite{EES} propose to learn a linear classifier per exemplar in the training set. The algorithm benefits from a large collection of simpler exemplar classifiers. In this way, the method is tuned to the appearance of the exemplar. While the detection of this detector covers the objects in the dataset (high recall), the detector usually provides low average precision. This is due to the large number of false detections introduced by each of the exemplar specific classifiers. Currently, remarkable results for object detection are obtained by convolutional neural networks~\cite{Overfeat,RCNN}. Girshick et al.~\cite{RCNN} employ the CNN of~\cite{ImageNet} to a set of candidate windows obtained by selective search~\cite{SSER}.

\subsection{Contextual Information for Object Detection}
Contextual information for object detection has been exploited over the past few years. Contextual information includes the relation between objects~\cite{Context1,Context10}, scene layout~\cite{Desai} or characteristics~\cite{hrContext,ContextPrime}, surrounding pixels~\cite{Context1,Context4,Context11} and background segments~\cite{Context3}. \cite{ContextPrime} shows that real-world scene structures can be modeled by inference rules. Therefore, in addition to the appearance of objects, contextual information provides useful information for object detection ~\cite{Survey,Survey2}. For example, Choi et al.~\cite{hrContext} model the object spatial relationships and co-occurrences by employing a tree-structured graphical model. Desai et al.~\cite{Desai} model the spatial arrangements between objects to detect objects in a structured prediction framework. Cinbis and Sclaroff~\cite{Cinbis} formulate the object and scene context in terms of relative spatial locations and relative scores between pairs of detections as sets of unordered items. Felzenszwalb et al.~\cite{DPM} re-score their DPM detections by exploiting contextual information as a post processing. Their re-scoring scheme relies on object co-occurrences as well as the location and size of the objects. The above methods show that contextual information is important for object detection. However, these methods have also certain limitations. For example, the above methods rely on object-object co-occurrences and spatial relationships and hence are suited for images consisting of (many) different objects. Further, the context-based methods aim at re-scoring detections. They do not introduce new detections and hence are not able to recover from missed detections of single detectors.

\subsection{Score Aggregation}
The approach of aggregating the responses of classifiers and learning a second level SVM to re-score them for different tasks such as action recognition~\cite{ICCV11}, image retrieval~\cite{Douze} and object recognition~\cite{Torresan,Context8} has been exploited in the literature. The organizers of Pascal VOC12 use seven methods submitted to the classification challenge. The scores of each submission are concatenated to form a single vector to train another linear classifier. Substantial increase for average precision is reported for classes such as potted plants and bottles. However, the problem of aggregating scores of different object detectors is not straightforward as other problems mentioned. More precisely, for these problems each instance in the dataset has a response from each classifier. By contrast, the object detectors do not generate candidate regions (exactly) at the same locations. Therefore, each candidate region does not necessarily has response from other detectors. Recently, Xu et al.~\cite{Pedest} propose to combine different pedestrian detectors using a score calibration and detection clustering steps. The authors reduce false and missed detections of pedestrian detectors per image. However, they do not aim at performing a global ranking of detections over all dataset for different object classes.


Our contributions are the following:

\begin{itemize}
  \item Detector combination: We propose to combine the state-of-the-art object detectors rather than proposing a new one.
  \item Detector consistency: We show that the state-of-the-art detectors have many detections in common. These common detections are proven to be very informative to re-rank detection scores.
  \item Detector complement: We show that existing state-of-the-art object detectors also have complementary detections. These complementary detections reduce missed detections of single detectors in a combined list.
  \item Detector contextual integration: We propose high-level context features (e.g. detector-detector relations and object-saliency cues) to combine detections in a learning to rank framework.
\end{itemize}

\section{Object Detectors}
In this section, the detectors used in this paper are outlined. We focus on publicly available detectors. Note that there are no constraints on the type of detector since the proposed method only requires detections (bounding box locations with classifier scores) of a detector.

\subsection{DPM}
Felzenszwalb et al.~\cite{DPM} propose an object detector in which each object category consists of a global template and deformable parts. The global template and deformable parts are represented by HOG features extracted at different scales. Training the object models is done in a latent support vector machines (SVM) framework. Each detection $\{x_1 , x_2, ... , x_n \}$ in training set is labeled $y_i$ as being either $+1$ or $-1$. Each detection $x$ is scored as

\begin{equation}
f_{\beta}(x) = \underset{z \in Z(x)} {\mathrm{max}} \beta . \Phi(x,z).
\label{eq:L-SVM}
\end{equation}

The set $Z(x)$ defines all possible latent values for detection $x$. $\beta$ and $\Phi(x,z)$ is a vector of model parameters and a feature vector, respectively. $\beta$ is trained by minimizing the following objective function:

\begin{equation}
L(\beta) = \frac{1}{2} ||\beta||^2 + C \sum_{i=1}^{n} max(0,1 -  y_if_{\beta}(x_i)),
\label{eq:objective}
\end{equation}

where $max(0,1 -  y_if_{\beta}(x_i))$ is the hinge loss and constant $C$ is the regularization parameter.

\subsection{CN}
Khan et al.~\cite{CNHOG} propose an object detector which uses color attributes as a complementary feature to DPM based HOG features. Color attributes are combined with HOG features in a late fusion manner. The proposed color attributes are compact and efficient. They are proven to be effective for the object classes in which intra class color variations are low such as potted-plants and sheep. Beside HOG features are extended with color attributes, training is done exactly the same as in DPM.

\subsection{EES}
Malisiewicz et al.~\cite{EES} propose an object detector which is trained by a parametric SVM for each positive exemplar in the training set. Consequently, a large collection of simpler exemplar specific detectors, which are highly tuned to the appearance of the exemplars, are obtained. Each exemplar is represented using a rigid HOG template~\cite{HOG} to train a linear SVM. Then, each Exemplar-SVM, ($\beta_E$,$b_E$), is used as a learned instance-specific HOG weight $\beta_E$ vector to score. $\beta_E$ is learned by optimizing the following convex objective function:

\begin{equation}
\Omega_E(\beta,b) = ||\beta_E||^2 + C_1h(\beta^Tx_E+b) + C_2 \sum_{x \in N_E} h(-\beta^Tx-b),
\label{eq:objectiveEES}
\end{equation}

where $h(x)=max(0,1-x)$ is the hinge loss and $C_1$ and $C_2$ are regularization parameters. Training each detector allows to tune detectors based on variations on the exemplar's appearance (viewpoint and object geometry). As a result, high recall is obtained for object detection.

\section{Combining Detectors by Learning to Rank\label{methodology}}

\begin{figure}[t]
\begin{center}
   \includegraphics[width=1\linewidth]{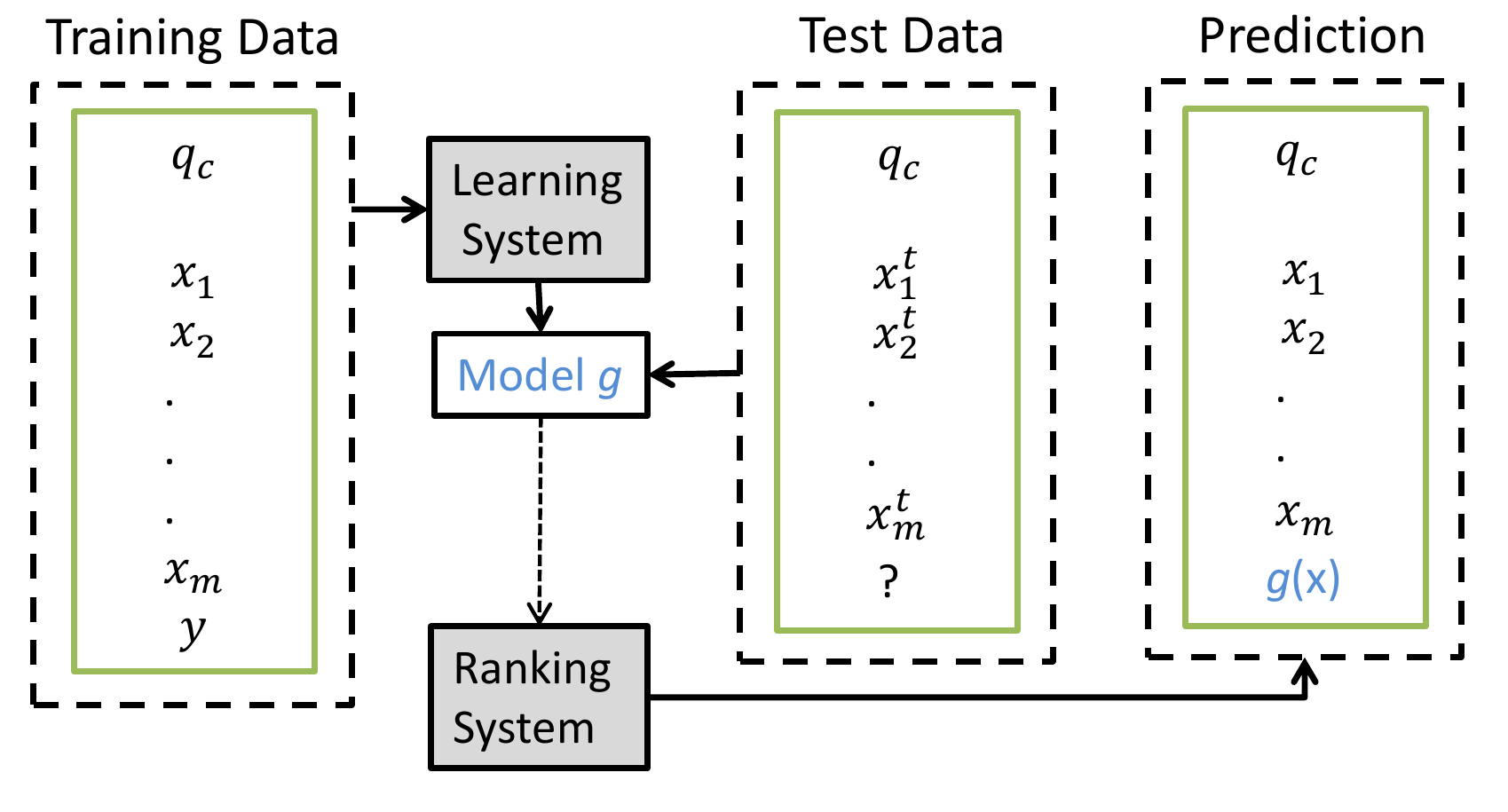}
\end{center}
   \caption{Learning to rank framework for detection re-ranking.}
\label{fig:L2R}
\end{figure}

To combine detections from different detectors, L2R is used. L2R aims at ranking group of items according to their relevance to a given task. Fig.~\ref{fig:L2R} illustrates a common L2R flow. In our framework, the training set consists of detections $X = \{x_i\}_{i=1}^{m}$ ($m$ is the number of the items in training set) and the ground truth label ($y$). Feature vector $\Phi$ and $y$ are used in training data to learn a ranking model ($g$). To re-score detections, $g$ is described as follows:

\begin{equation}
g(x) = w\Phi(x).
\label{eq:model}
\end{equation}

Using different loss functions $\xi$ (see section~\ref{experiment}), the weight ($w$) is optimized by minimizing the following objective function:

\begin{equation}
\min_w \frac{1}{2} w^{T}w + C \sum_{i=1}^{l} \xi_i.
\label{eq:objective}
\end{equation}


To learn a ranking algorithm to perform re-ranking, the proposed method starts with the feature extraction step using detections $x$ from different detectors.

\begin{figure*}
\begin{center}
   \includegraphics[width=1\linewidth]{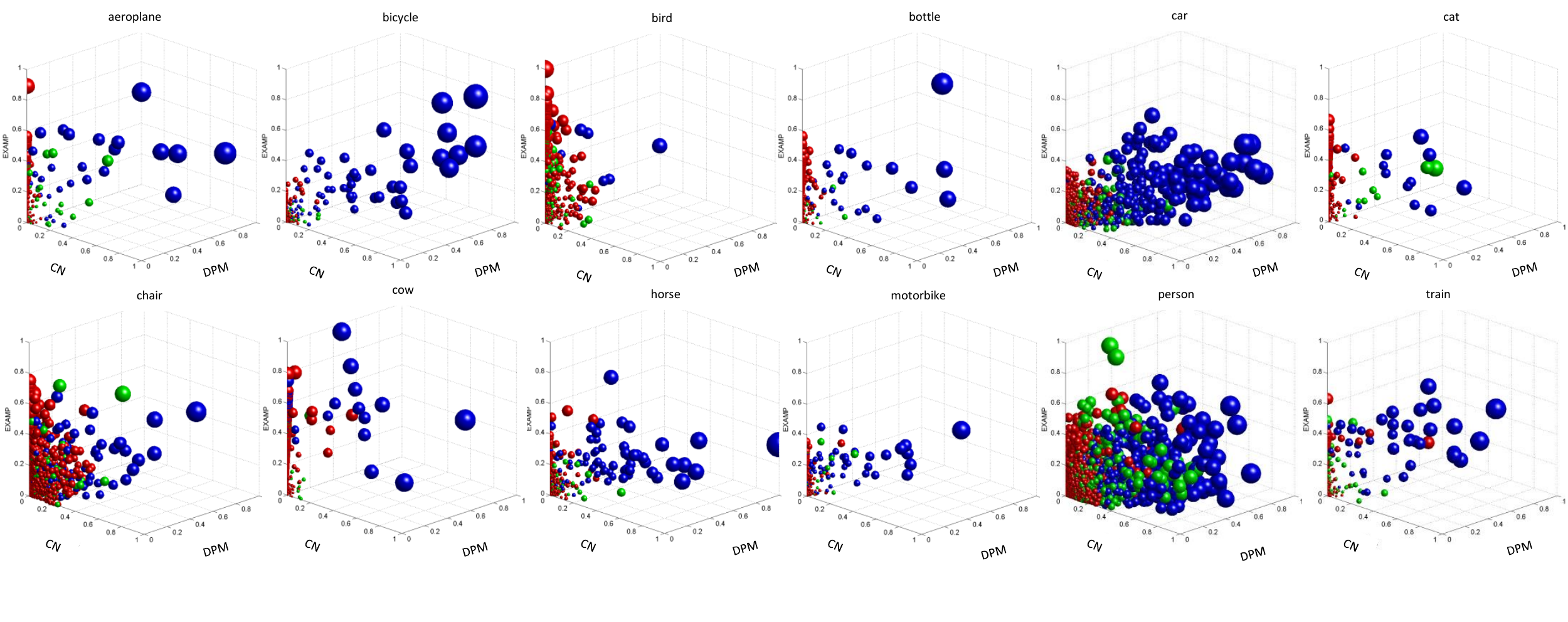}
\end{center}
   \caption{The figure illustrates relative score $R$ for each detection in VOC 2007 trainval set. Each sphere represents a detection in the trainval set whereas each axis represents relative score from detectors namely, DPM, CN and EES. The color blue, green and red holds for true detection, poor localization and false detection, respectively. Best viewed in color.}
\label{fig:Relative}
\end{figure*}

\subsection{Context Features}

The proposed method starts with high-level context feature extraction to learn the ranking between detections of different detectors into a single detection list. We aim at extracting generic features exploiting the correlation and consistency between detectors.

\subsubsection{Detector-Detector Context}


We define detector consistency when different object detectors generate detections for the same image region. Agreement of all object detectors for a certain location increases the probability of a correct object detection. However, different detectors may generate detections at different locations even for the same image. As a result, it is hard to obtain an exact bounding box location where all detectors provide a detection. Therefore, a detector relative score is defined. To obtain a relative score for each detection, a correspondence term is computed by considering the overlapping ratios between all other detections. In this way, an image is represented as a collection of detections obtained by different object detectors $j$, where $j=\{1, 2 \dots n \}$ and $n$ is the number of the detectors used. For the $i^{th}$ detection in the image, the maximum overlapping detection with each detector is considered as follows:

\begin{align}
A_{i,j} = \frac{Area(BB_i \cap BB_j)}{Area(BB_i \cup BB_j)} \label{eq:detcor}\ \ ,\\
[\Gamma_i(j),\varphi_i(j)] = \max{(A_{i,j})} \ \ ,
\label{eq:detcorr}
\end{align}

where $\Gamma$ is the overlap ratio and $\varphi$ is the index of the maximum overlapping detection for detector type $j$. Then, the corresponding relative score $R$ of a detector $j$ to the $i^{th}$ detection is considered as $R_{i,j} = \Gamma_i(j) \times S(\varphi_i(j))$, where $S$ is the initial classification score of the detector. Note that if a detection has no overlap with other detectors ($\Gamma_i(j)=0$), its relative scores will be zero. In this way, higher relative scores correspond to more reliable detections because more detectors agree on a particular location (see Fig.~\ref{fig:Relative}). If a detection has high relative score from each single detector it corresponds to a high probability of being a true detection. Whereas a low relative score corresponds to a false detection. Moreover, a mid-level consistency in relative score can be considered as a good indication of poor localization error.

Relative score of a detection does not include the information of which detector it belongs to. However, some detectors performs better than others for some classes, hence its detections should get higher scores than detection of other detectors (to benefit the strength of detectors on the task they are successful). Therefore, the detector indicator term is specified. The aim is to provide information to the learning system to create detector preferences over classes. To give an indication of which detector the detection belongs to, a binary vector $I_D$ of three dimensions (i.e. three detectors in our case) is used. The value of the dimension is assigned to be one if case of a detection by the corresponding detector otherwise the value is set to zero. This feature vector is at the detector level. Therefore, all detections of the same detector have the same binary coding $I_D$.

The final corresponding score feature $Rs$, for the $i^{th}$ detection is denoted by $Rs_i =$ $\{ I_{D,i},$ $ R_{i,1},$ $R_{i,2},...,$ $R_{i,n}, R_{i,1}+R_{i,2},R_{i,1}+R_{i,3}, R_{i,2}+R_{i,3}, ...,R_{i,n-1}+R_{i,n}, R_{i,1}+R_{i,2}+R_{i,3}+...,+R_{i,n} \}$. The dimension of $Rs$ is limited to the number of the detectors.

%

\begin{figure}
\begin{center}
   \includegraphics[width=.7\linewidth]{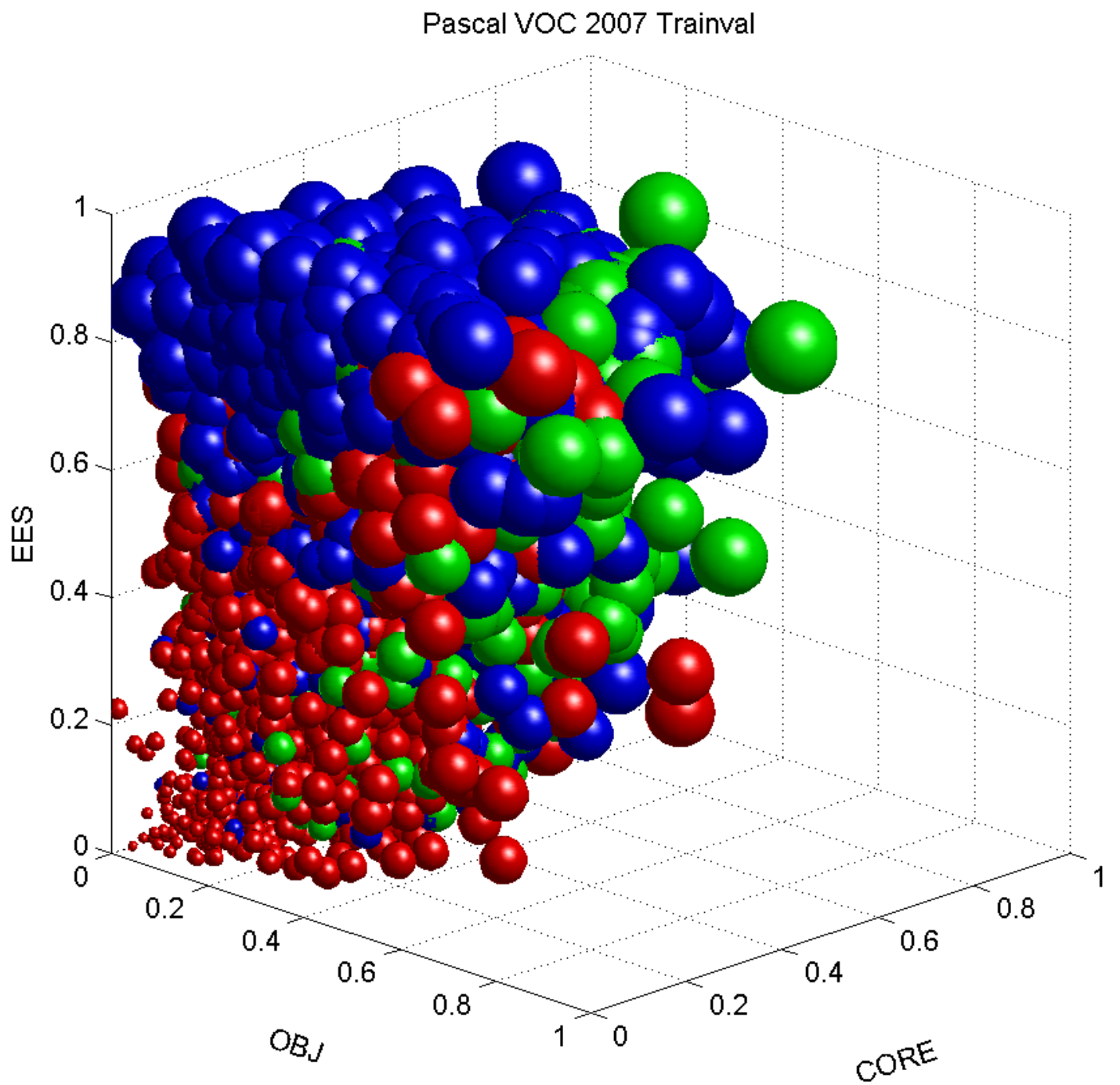}
\end{center}
   \caption{The figure illustrates object likelihood score $O_s$ for each detection in VOC 2007 trainval set. Each sphere represents a detection (randomly sub-sampled over all classes) in the trainval set whereas each axis represents object likelihood score from object indicators namely, OBJ, CORE and EES. The color blue, green and red holds for true detection, poor localization and false detection, respectively. Best viewed in color.}
\label{fig:ObjLikelihood}
\end{figure}

\subsubsection{Object-Saliency}
\vspace{5px}
A feature vector $Os$ is proposed to represent how likely a detection contains an object. EES~\cite{EES}, OBJ~\cite{Objectness} and CORE~\cite{CORE} are used to measure the object-saliency of a detection. OBJ and CORE are category independent region proposal methods. They are mostly used by the current object detection algorithms to avoid exhaustive sliding window search. These methods provide region candidates/proposals (bounding box) which most likely contain objects. Both methods approximately result in 1000 region candidates per image. In addition to these category independent region proposal methods, EES~\cite{EES} is also used to provide region candidates. The overlap ratios between these different region proposals and object detections are calculated according to eq.~\ref{eq:detcor}. Then, the feature vector $Os$ for the $i^{th}$ detection is given by:


\begin{align}
\Psi_{i,j} = sort(A_{i,j}) \ \ ,\\
Os_{i,j} = \frac{1}{n} \sum \limits_{k=1}^n \Psi_{i,j}(k),
\label{eq:Objectness2}
\end{align}

where $n$ is the number of neighbors to measure object-saliency, $\Psi$ is the sorted list of overlaps and $j$ is the indicator of different regions proposals, namely OBJ, CORE and EES. Additionally, we use the confidence scores of the maximum overlapping neighbors of detections by EES~\cite{EES} in eq.~\ref{eq:Objectness2} since these regions proposals are class specific. A detection with a high object-saliency value is considered to be a good indicator for a correct detection. These features may be useful to provide less confidence scores for false detections. Fig.~\ref{fig:ObjLikelihood} illustrates that true or false detections are highly correlated with object likelihood scores of the detection.

\subsubsection{Object-Object Relation}
\vspace{5px}
The likelihood of an object present is inferred by using other object class likelihoods.  Let $S_{c,j}$ be the detection with maximum confidence for object class $c$ ($c=\{1,2,\dots ,m \}$) by detector $j$ (j=\{1,2,3\}) in an image, where $m$ denotes the number of object classes. Then, the object-object context $So$ is given by

\begin{align}
So(c) = \sum\limits_{j=1}^3 {S_{c,j}}.
\label{eq:OOR}
\end{align}

This feature exploits the object-object relations. For instance, when three detectors locate a cow with high confidence, it is less likely to have a sofa or tv in the same image.

The compactness of the proposed contextual features used in this paper is shown in Table~\ref{table:Feature}.


\begin{table}
\begin{center}
\renewcommand\arraystretch{1.2}
\scalebox{1.2}{
\begin{tabular}{ l  c  c }
\midrule[1.6pt]
\noalign{\vskip.2pt}
                    Feature & Notation & Dimension\\
\noalign{\vskip.2pt}
\midrule[1.6pt]
\noalign{\vskip.2pt}
\multirow{4}{*}{}
Detector Relative Score    & $Rs$  & 10\\
Object Likelihood Measure     & $Os$  & 4\\
Object-Object Context   & $So$  & 20 \\ \noalign{\vskip.2pt} \hline \noalign{\vskip.2pt}
Total                  &       & 34 \\
\noalign{\vskip.2pt}
\midrule[0.5pt]
\noalign{\vskip.2pt}

\end{tabular}
}
\end{center}
\caption{\label{table:Feature} Contextual features used in the proposed learning to rank framework.}
\end{table}

\subsection{Learning}\label{learning}
\begin{table*}[t]
\begin{center}
\renewcommand\arraystretch{1.2}
\scalebox{0.81}{
\begin{tabular}{| l  l | c | c | c | c | c | c | c | c | c | c | c | c | c | c | c | c | c | c | c | c | c |}
\midrule[1.6pt]
\noalign{\vskip.2pt}
                     &  & aero & bike & bird & boa & bot & bus & car & cat & chr & cow & tab & dog & hor & mbik & pers & plnt & shp & sofa & tra & tv & mAP\\
\noalign{\vskip.2pt}
\midrule[1.6pt]
\noalign{\vskip.2pt}
\multirow{10}{*}{}
&DPM~\cite{DPM}  & 	26.7 	 & 	56.9 	 & 	2.6 	 & 	12.8 	 & 	21.9 	 & 	46.0 	 & 	55.3 	 & 	13.7 	 & 	 19.0 	 & 	19.4 	 & 	12.6 	 & 	 2.2 	 & 	58.1 	 & 	47.3 	 & 	 40.9 	 & 	6.8 	 & 	15.0 	 & 	 26.9 	 & 	 43.4 	 & 	38.8 	 & 	28.3 \\ \hline
&CN~\cite{CNHOG} & 	28.7 	 & 	55.9 	 & 	6.3 	 & 	11.6 	 & 	18.2 	 & 	44.3 	 & 	55.5 	 & 	17.7 	 & 	 18.3 	 & 	20.5 	 & 	14.9 	 & 	 4.9 	 & 	57.3 	 & 	48.9 	 & 	 41.5 	 & 	15.0 	 & 	21.8 	 & 	 28.1 	 & 	 44.1 	 & 	45.7 	 & 	30.0 \\ \hline
&EES~\cite{EES}  & 	17.9 	 & 	47.2 	 & 	2.8 	 & 	10.6 	 & 	9.1 	 & 	39.3 	 & 	40.3 	 & 	1.6 	 & 	 6.2 	 & 	15.3 	 & 	7.0 	 & 	 1.7 	 & 	44.0 	 & 	38.1 	 & 	 13.2 	 & 	4.6 	 & 	20.0 	 & 	 11.6 	 & 	 35.9 	 & 	27.6 	 & 	19.7 \\ \hline
\midrule[1.6pt]
&M-DPM     & 	39.3 & 	66.5 & 	29.2  	 & 	25.5 	 & 	36.2 	 & 	58.2 	 & 	73.4 	 & 	36.3 	 & 	53.8 	 & 	 33.6 	 & 	19.9 	 & 	 22.5 	 & 	74.7 	 & 	65.5 	 & 	62.5 	 & 	 35.0 	 & 	28.5 	 & 	37.2 	 & 	 66.0 	 & 	 51.6 	 & 	45.8 \\ \hline
&M-CN   & 	43.5 	 & 	61.7	 & 	26.1  	 & 	20.5 	 & 	34.5 	 & 	56.8 	 & 	72.2 	 & 	39.4 	 & 	 46.3 	 & 	 33.2 	 & 	22.8 	 & 	22.5 	 & 	73.3 	 & 	63.1 	 & 	65.0 	 & 	 38.3 	 & 	38.8 	 & 	 43.9 	 & 	62.1 	 & 	 60.1 	 & 	46.2   \\ \hline
&M-EES     & 	47.7 & 	72.4 	 & 	38.3     & 	37.3 	 & 	46.1 	 & 	64.3 	 & 	64.1 	 & 	45.0 	 & 	 44.4 	 & 	 50.8 	 & 	44.7 	 & 	 43.1 	 & 	69.5 	 & 	63.4 	 & 	54.9 	 & 	 35.6 	 & 	47.9 	 & 	 50.2 	 & 	62.8 	 & 	 73.7 	 & 	\bf{52.8} \\ \hline
\midrule[1.6pt]
&M-(DPM + EES)     & 	60.7 	 & 	80.4     & 	48.6 	  	 & 	46.0 	 & 	54.6 	 & 	73.7 	 & 	80.2 	 & 	 59.5 	 & 	67.6 	 & 	58.2 	 & 	51.9 	 & 	51.3 	 & 	82.2 	 & 	 73.8 	 & 	73.1 	 & 	49.0 	 & 	 51.7 	 & 	60.3 	 & 	76.2 	 & 	76.0 	 & 	63.7\\ \hline
&M-(DPM + CN)   & 	48.8 	& 	68.0 	 & 	36.8   & 	27.4 	 & 	40.7 	 & 	62.9 	 & 	77.4 	 & 	49.4 	 & 	 61.5 	 & 	39.8 	 & 	31.6 	 & 	33.7 	 & 	78.7 	 & 	70.8 	 & 	 71.2 	 & 	48.5 	 & 	42.1 	 & 	 49.4 	 & 	 70.6 	 & 	61.7 	 & 	53.5 \\ \hline
&M-(EES + CN)  & 	59.3 	 & 	79.5  & 	46.6 	  & 	43.7 	 & 	54.6 	 & 	72.8 	 & 	78.9 	 & 	 62.0 	 & 	 63.2 	 & 	55.7 	 & 	51.5 	 & 	52.1 	 & 	82.5 	 & 	71.7 	 & 	 74.7 	 & 	50.0 	 & 	 55.8 	 & 	66.1 	 & 	 73.4 	 & 	76.3 	 & 	63.5 \\ \hline
\midrule[1.6pt]
&M-All    & 	62.5 	 & 	81.3  & 	52.3  	 & 	47.5 	 & 	56.7 	 & 	76.1 	 & 	82.3 	 & 	65.9 	 & 	 71.2 	 & 	59.0 	 & 	55.3 	 & 	 56.9 	 & 	84.2 	 & 	75.7 	 & 	 77.5 	 & 	56.0 	 & 	57.0 	 & 	 67.4 	 & 	78.4 	 & 	76.3 	 & 	\bf{67.0} 	 \\ \hline
\noalign{\vskip.2pt}
\end{tabular}
}
\end{center}
\caption{\label{table:Bound}  mAP values for baseline detectors DPM, CN and EES. Class specific and overall maximal mAP values of baseline detectors M-DPM, M-CN and M-EES, and their combinations M-(DPM+CN), M-(CN+EES), M-(DPM+EES) and M-(All) on PASCAL VOC07.}
\end{table*}


L2R methods are used to learn the ranking models. L2R methods used in this paper can be categorized in two groups~\cite{L2R}. The first type of algorithms is called pointwise techniques. Pointwise approaches represent the problem of ranking as a regression or classification problem. These techniques are straightforward approaches to learn the ranking model. Pointwise algorithms are preferred because of their efficiency and effectiveness. These methods have been optimized to work on large scale data.

The second type of L2R algorithms are pairwise techniques. These methods consider the problem of ranking as a pairwise classification problem. The aim is to learn a binary classifier to determine which instance is most relevant from a given pair of instances. The goal of these algorithms is to minimize the average number of miss orders in ranking rather than the traditional miss classification in the ordinary pointwise approach.


\section{Non-maximum Suppression}\label{nms}
Duplicate removal for the same instance is a known problem for single detectors. Obviously, by combining multiple detectors, the proposed method increases the number of duplicates. To this end, we propose to eliminate these multiple detections by non-maximum suppression ($nms$). The common application of $nms$ considers all bounding boxes (over a certain overlap threshold) for suppression. We use only correspondences (overlaps between detections of other detectors) obtained for each detection in eq.~\ref{eq:detcorr} for suppression. After applying the re-ranking system, the corresponding detections are sorted and the highest among the others is remained constant while the other detections which are at least $40\%$ covered by the highest detection are suppressed.

\section{Experiments}\label{experiment}

\begin{figure}
\begin{center}
   \includegraphics[width=.9\linewidth]{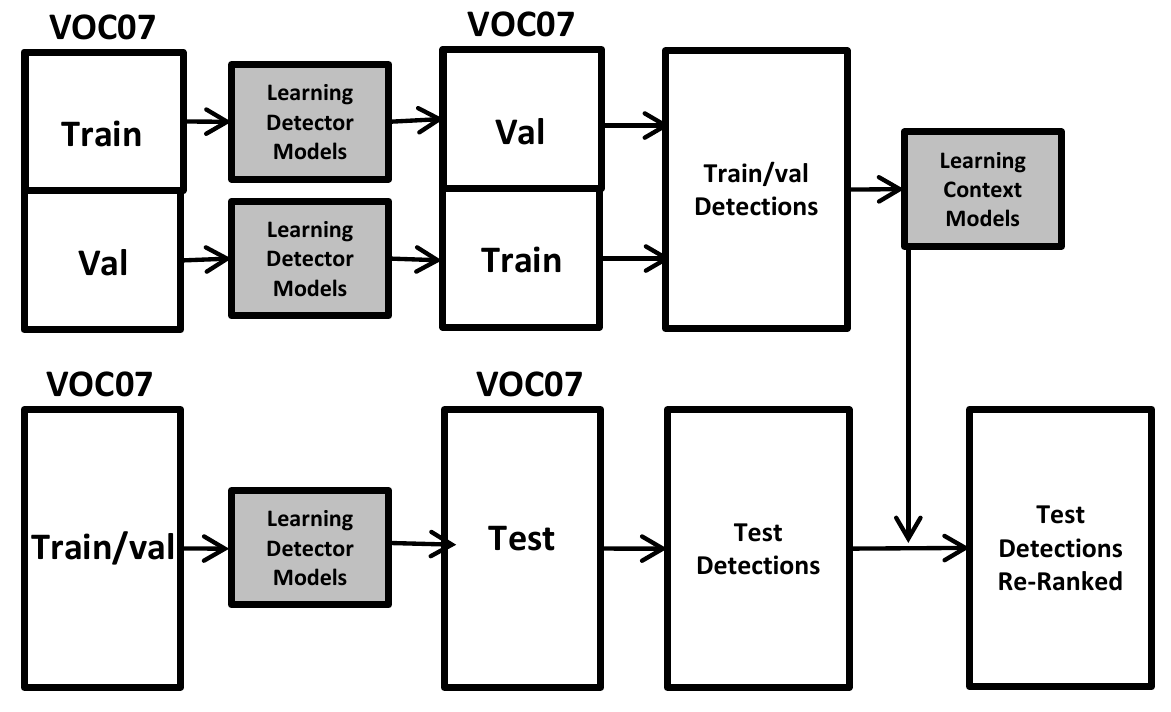}
\end{center}
   \caption{Each training is used for learning detector models and context models. To avoid overfitting, the object detectors for context models are trained on the train set to generate detections on validation. Further, they are trained on validation to provide detections on train.}
\label{fig:trainprocess}
\end{figure}

Experiments are conducted on the Pascal VOC07 and VOC10 datasets. VOC07 dataset consists of 9963 images of 20 different object classes (24640 annotated objects) with 5011 training images and 4952 test images. VOC10 train/val dataset contains 10103 images of 20 different categories (23374 annotated objects). Object detections for the $train$ set are obtained by models trained on 2007$val$ and detections for the $val$ set are trained on 2007$train$ set to learn detector-detector context. Detections for the $test$ set are obtained by models trained on the 2007$trainval$ set for both dataset evaluations. This process is summarized in Fig.~\ref{fig:trainprocess}.

\begin{table*}[t]
\begin{center}
\renewcommand\arraystretch{1.2}
\scalebox{0.85}{
\begin{tabular}{| l  c | c | c | c | c | c | c | c | c | c | c | c | c | c | c | c | c | c | c | c | c | c |}
\midrule[1.6pt]
\noalign{\vskip.2pt}
                     &  & aero & bike & bird & boa & bot & bus & car & cat & chr & cow & tab & dog & hor & mbik & pers & plnt & shp & sofa & tra & tv & mAP\\
\noalign{\vskip.2pt}
\midrule[1.6pt]
\noalign{\vskip.2pt}
\multirow{11}{*}{}
&DPM~\cite{DPM}  & 	26.7 	 & 	56.9 	 & 	2.6 	 & 	12.8 	 & 	21.9 	 & 	46.0 	 & 	55.3 	 & 	13.7 	 & 	 19.0 	 & 	19.4 	 & 	12.6 	 & 	 2.2 	 & 	58.1 	 & 	47.3 	 & 	 40.9 	 & 	6.8 	 & 	15.0 	 & 	 26.9 	 & 	 43.4 	 & 	38.8 	 & 	28.3 \\ \hline
&CN~\cite{CNHOG} & 	28.7 	 & 	55.9 	 & 	6.3 	 & 	11.6 	 & 	18.2 	 & 	44.3 	 & 	55.5 	 & 	17.7 	 & 	 18.3 	 & 	20.5 	 & 	14.9 	 & 	 4.9 	 & 	57.3 	 & 	48.9 	 & 	 41.5 	 & 	15.0 	 & 	21.8 	 & 	 28.1 	 & 	 44.1 	 & 	45.7 	 & 	30.0 \\ \hline
&EES~\cite{EES}  & 	17.9 	 & 	47.2 	 & 	2.8 	 & 	10.6 	 & 	9.1 	 & 	39.3 	 & 	40.3 	 & 	1.6 	 & 	 6.2 	 & 	15.3 	 & 	7.0 	 & 	 1.7 	 & 	44.0 	 & 	38.1 	 & 	 13.2 	 & 	4.6 	 & 	20.0 	 & 	 11.6 	 & 	 35.9 	 & 	27.6 	 & 	19.7 \\ \hline
\midrule[1.6pt]
&NaiveI      & 	31.0 	 & 	61.6 	 & 	6.1 	 & 	13.7 	 & 	22.7 	 & 	48.9 	 & 	58.4 	 & 	19.6 	 & 	 20.5 	 & 	22.3 	 & 	19.3 	 & 	3.9 	 & 	63.2 	 & 	52.1 	 & 	 44.3 	 & 	14.5 	 & 	22.7 	 & 	 31.5 	 & 	47.8 	 & 	47.4 & 32.6\\ \hline
&NaiveII     & 	30.8 	 & 	57.7 	 & 	6.1 	 & 	14.2 	 & 	20.2 	 & 	47.6 	 & 	55.2 	 & 	13.3 	 & 	 16.7 	 & 	22.4 	 & 	20.0 	 & 	4.4 	 & 	61.4 	 & 	50.2 	 & 	 33.4 	 & 	11.8 	 & 	23.4 	 & 	 28.0 	 & 	46.4 	 & 	41.6 	 & 	30.2\\ \hline
&NaiveIII    & 	28.3 	 & 	61.3 	 & 	2.8 	 & 	13.3 	 & 	22.8 	 & 	48.1 	 & 	58.7 	 & 	18.5 	 & 	 19.5 	 & 	15.3 	 & 	19.0 	 & 	1.8 	 & 	61.7 	 & 	52.6 	 & 	 41.9 	 & 	14.8 	 & 	20.0 	 & 	 29.3 	 & 	48.9 	 & 	48.3 	 & 	31.4\\ \hline
\midrule[1.6pt]
& PoW1       & 	36.8 	 & 	62.7 	 & 	10.0 	 & 	18.1 	 & 	24.3 	 & 	51.6 	 & 	59.5 	 & 	21.2 	 & 	 22.5 	 & 	25.4 	 & 	22.4 	 & 	7.8 	 & 	64.2 	 & 	57.3 	 & 	 44.9 	 & 	18.7 	 & 	26.7 	 & 	 34.1 	 & 	54.1 	 & 	47.8 	 & 	35.5 \\ \hline
& PoW2       & 	36.7 	 & 	62.8 	 & 	13.3 	 & 	18.4 	 & 	27.0 	 & 	52.3 	 & 	59.9 	 & 	24.7 	 & 	 21.9 	 & 	24.8 	 & 	25.8 	 & 	10.6 	 & 	65.4 	 & 	55.9 	 & 	 44.7 	 & 	19.2 	 & 	21.2 	 & 	 37.5 	 & 	54.0 	 & 	46.5 	 & 	36.2\\ \hline
& PoW3       & 	35.6 	 & 	63.1 	 & 	9.7 	 & 	17.0 	 & 	25.0 	 & 	51.2 	 & 	60.0 	 & 	21.3 	 & 	 22.5 	 & 	25.1 	 & 	21.5 	 & 	8.1 	 & 	65.0 	 & 	56.4 	 & 	 43.8 	 & 	18.2 	 & 	27.0 	 & 	 33.9 	 & 	53.5 	 & 	48.2 	 & 	35.3\\ \hline
& PaW1           & 34.5 & 59.4 & 10.2 & 16.2 & 19.8 & 49.5 & 54.4 & 24.6 & 20.7 & 19.7 & 24.0 & 8.0 & 61.0 & 51.5 & 40.9 & 16.7 & 25.9 & 31.1 & 48.3 & 41.5 & 32.9\\
\midrule[1.6pt]
&Imp             & 	8.1 	 & 	7.1 	 & 	6.2 	 & 	5.6 	 & 	5.1 	 & 	6.3 	 & 	4.5 	 & 	7.0 	 & 	 3.5 	 & 	4.9 	 & 	10.9 	 & 	5.7 	 & 	7.4 	 & 	8.4 	 & 	3.4 	 & 	4.2 	 & 	5.2 	 & 	 9.4 	 & 	 9.9 	 & 	2.4 	 & 6.3\\
\noalign{\vskip.2pt}
\midrule[1pt]
\noalign{\vskip.2pt}
\end{tabular}
}
\end{center}
\caption{\label{table:Result} The results using learning to rank algorithms. Naive: Direct merging methods without learning. Imp: The improvement over maximum baseline detector by maximum learning algorithm.}
\end{table*}

\subsection{Detector Bounds}

In this experiment, we evaluate the maximal mAP that can be achieved by the detections of the baseline detectors and their combinations. The maximal mAP of a detector is calculated when all true detections are ranked at the top of the detection list. Table~\ref{table:Bound} shows that re-ranking $DPM$, $CN$ and $EES$ detections results in a substantial performance improvement, $17.5\%$, $16.2\%$ and $33.1\%$, respectively.  This result shows the positive effect on re-ranking detection scores of object detectors.

Table~\ref{table:Bound} shows that $DPM$ and $CN$ have similar maximal mAP $45.6\%$ and $46.2\%$, respectively. However, their combination has significantly higher maximal mAP ($53.5\%$) than both of them individually. This shows that although these two detectors are very similar in nature, they are somewhat complementary to each other. Furthermore, when the detectors are designed intrinsically different (e.g. $DPM$ and $EES$ or $CN$ and $EES$), they are more complementary to each other. This can be derived by the performance gain obtained by combining DPM+EES and CN+EES in Table~\ref{table:Bound}, $10.9\%$ and $10.7\%$, respectively. Consequently, the proposed method would benefit from more detectors.

Another observation that can be derived from Table~\ref{table:Bound} is that beside detectors have complementary detections to each other, they also have common detections. While these common detections are useful to learn consistency in their output, complementary detections resolves missed detections for each individual detector.

Table~\ref{table:Bound} shows that the performance of detectors is limited by their correct detections. Therefore, detector combinations always show higher mAP values than each single detector. The proposed method highly benefits from this, whereas other context based re-ranking methods lead to a limited performance improvement (limited to correct detections of a single detector).

\subsection{Direct Combining of Detections}

In this experiment, several ways of combining (without learning) detector outputs are investigated. Since the detectors are trained independently, detector scores are not necessarily compatible. A calibration process~\cite{Platt} is applied before merging different detector outputs. Given a detection $x$ and the learned sigmoid parameter ($\alpha$, $\beta$), the calibrated detection score is calculated as

\begin{align}
f(x|\alpha,\beta) = \frac{1}{1+exp(x\alpha + \beta)},
\label{eq:DC}
\end{align}

where $\alpha$ and $\beta$ for each detector are learned in $trainval$ set. After the scores are calibrated, we evaluate three different approaches of combining:

\begin{itemize}
  \item $\bf{NaiveI}$, after scores are calibrated, detections are merged into a single list.
  \item $\bf{NaiveII}$, after scores are calibrated, detections are sorted in a descending score order for each single detector. Then, detections are combined by taking one by one from the top of each sorted detector outputs.
  \item $\bf{NaiveIII}$, the detectors are combined based on their training set performance. The output of the best performing detector is first added to the list followed by the others based on their performance.
\end{itemize}

After the detections are combined in a single list, $nms$ (see section~\ref{nms}) is applied. It can be derived from Table~\ref{table:Result} that naively combining detector outputs outperforms baseline scores. The improvements are due to the increase in recall of the combined detection list.

The minimum performance improvement is obtained by $NaiveII$. $NaiveII$ gives equal importance to each single detector. This means that although $EES$ detections are not precise, they become as important as $DPM$ and $CN$. Therefore, more false positives are introduced at the top of detection list which negatively affects the detection performance. This result shows the importance of properly weighting the detections.

$NaiveIII$ is expected to perform better than other naive methods since it incorporates the training performances of the baseline detectors. However, the $trainval$ performance of the baseline detectors explains the lower performance of $NaiveIII$. To obtain $trainval$ performance detector models are: $a)$ trained on $train$ to test on $val$ and $b)$ trained on $val$ to test on $train$ (see Fig.~\ref{fig:trainprocess}). Since the detectors are trained with fewer samples for $trainval$ detections, baseline performances do not necessarily correspond to their $test$ performances. Training with fewer examples has also an influence on our context models.

\subsection{Learning to Rank Detectors}

In this experiment, four different L2R algorithms are evaluated. Pointwise methods are $L2$ regularized support vector classifier ($PoW1$), logistic regressor ($PoW2$) and support vector regressor ($PoW3$). $RankSVM$~\cite{rankSVM} is commonly used as a pairwise L2R method. Therefore, $RankSVM$ ($PaW1$) is used as a pairwise method in our experiments. $Liblinear$~\cite{liblinear} implementations for pointwise approaches and rankSVM implementation by Joachims~\cite{rankSVM} are used with default parameter settings.

Ground-truth overlap ratios are taken as training labels. Pascal VOC ($>0.5$) overlap criteria is used to assign positive and negative labels for $PoW1$ and $PoW2$, while overlap ratios are directly used as training labels for $PoW3$ and $PaW1$.

Table~\ref{table:Result} shows that the proposed learning to rank approach outperforms the baseline detectors for all classes, $DPM$(7.8\%), $CN$(6.2\%) and $EES$(16.5\%). While learning based methods always perform better, logistic regression ($PoW2$) based learning method performs slightly better than other L2R algorithms. The performance of $RankSVM$ is slightly lower than other L2R methods. This is due to unbalanced data. The number of negative samples is significantly larger than positive samples. This has also influence on the final result of $RankSVM$.

Considering the low dimensionality of the proposed feature vector, the feature space may not be linearly separable. Therefore, other non-linear kernel options for classifier could be tried. However, we avoid learning a non-linear $SVM$ due to its learning time and parameter selection. Therefore, we use a feature mapping method proposed by Vedaldi and Zisserman~\cite{FeatMap}. A $34$ dimensional feature vector is mapped to a higher dimensional feature space. The best performing linear classifier in Table~\ref{table:Result}($PoW2$) is selected to perform on this new feature space. $PoW2$ classifier obtains $0.6\%$ mAP improvement($36.8\%$). Increasing the dimensionality results in support vectors to better separate the feature space. Increasing the feature vector dimension with additional context features may further improve the results.

The improvement over direct merging methods by the proposed learning scheme in Table~\ref{table:Result} indicates that the performance gain is not only due to the recall increase but also the effectiveness of the contextual information and learning scheme.

\subsection{Detection Error Analysis}\label{Section:ErrorAnalysis}

\begin{figure}[t]
\begin{center}
   \includegraphics[width=1\linewidth]{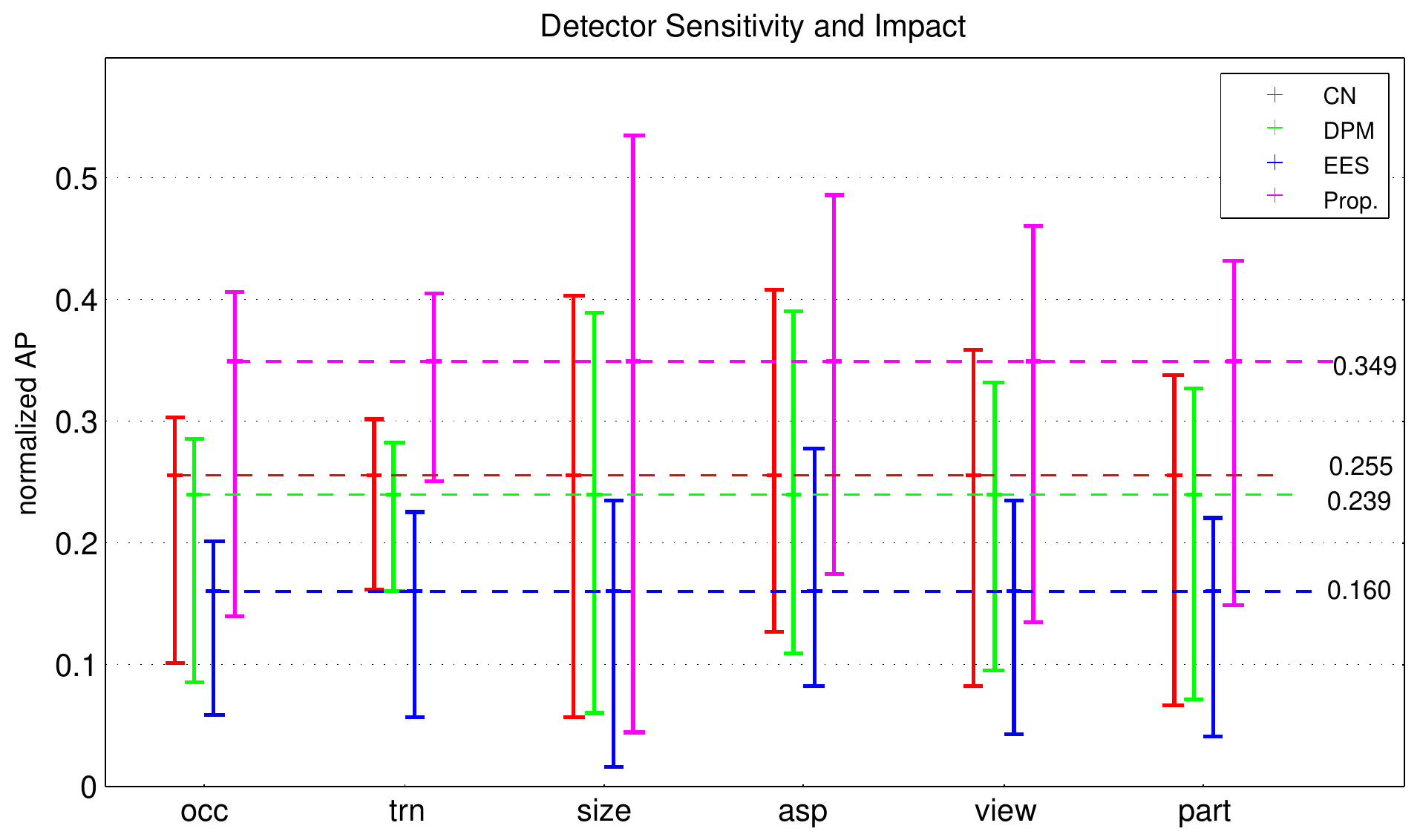}
\end{center}
   \caption{Average (over classes) $AP_N$ for the highest and lowest performing subsets within each different object characteristics such as occlusion, truncation, bounding box area, aspect ratio, viewpoint and part visibility.}
\label{fig:sens}
\end{figure}

To provide more insight in the performance obtained by combining the baseline detectors, we follow the procedure introduced by Hoiem et al.~\cite{Diagnose}. The first analysis is performed for detector sensitivities. The detector sensitivity is calculated based on the difference between max and min normalized AP for each characteristic (occlusion, truncation, bounding box area, aspect ratio, viewpoint, part visibility). Each different colored plot in Fig.~\ref{fig:sens} shows the mean (over all classes) normalized AP for specified detectors. The results show that the proposed method does not reduce the sensitivity. However, it improves both the highest and lowest performing subsets for nearly all object characteristics. This indicates that the proposed method improves robustness for all object characteristics. The reason why the proposed method does not reduce the sensitivity is due to commonly missed detections (hard detections cannot be detected easily even for human observers). While some of these hard detections are covered by one of the baseline detectors, they mainly remained unveiled. That is why the minimum normalized APs for each characteristic increase but not as much as the maximum normalized APs. Consequently, the difference between max and min normalized AP increases.

Hoiem et al.~\cite{Diagnose} show the problem of small objects. Since small sized objects are mainly missed by all detectors, even if three baseline detectors are combined, we observe that the min normalized AP for category "size" is not improved.

\begin{figure}[h]
\begin{center}
   \includegraphics[width=.9\linewidth]{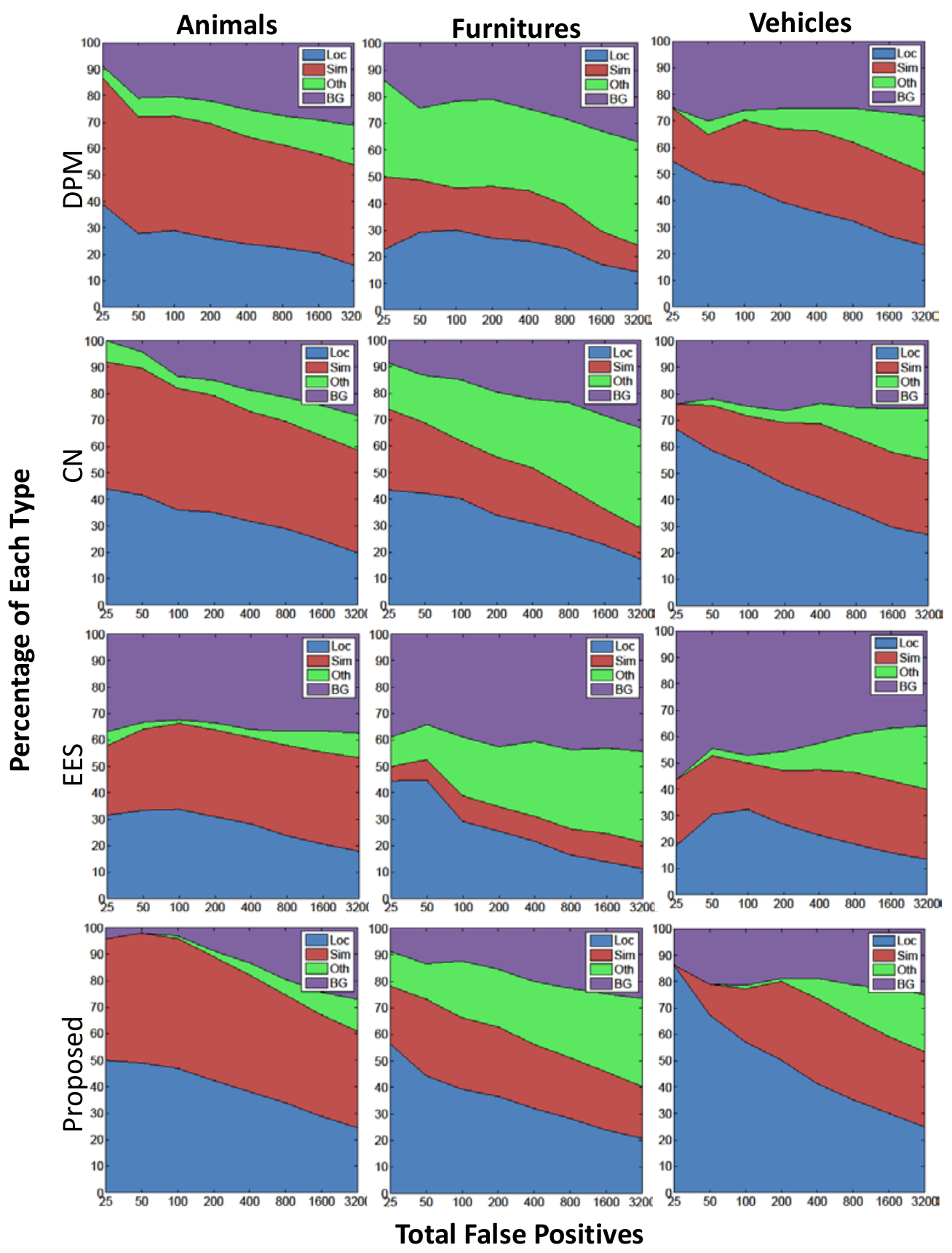}
\end{center}
   \caption{Figure shows the fraction of false positives of each type (animal, furniture and vehicle) evolving as the total number of false positives increase.}
\label{fig:FPA}
\end{figure}

Fig.~\ref{fig:FPA} shows the changes in the percentage of each false positive ($FP$) types with an increasing total number of $FP$. $FPs$ are divided into four categories as follows:

 \begin{itemize}
  \item Poor localization ($Loc$) occurs when the label of detection is correct but misaligned with the ground-truth detection ($0.1$ $\leq$ overlap $\leq$ $0.5$ or a duplicate detection.
  \item Confusion with similar classes ($Sim$) occurs when a false detection has an overlap with an instance of a similar class.
  \item Confusion with dissimilar object categories ($Oth$) occurs when a false detection is obtained for dissimilar classes.
  \item Confusion with background ($BG$) occurs when a false detection has no overlap with an instance of similar or dissimilar classes.
\end{itemize}

The obtained errors originate from poor localization rather than other errors. This shows the effectiveness of relative score features. For instance, consider an image region where all detectors generate a detection. All detections belonging to this region have high classifier scores because of the high relative score. Consequently, these detections are ranked at the top of the detection list. However, the proposed method creates preferences over detectors for classes. Now, assume a detection of preferred detector with a localization error in this region. The corresponding detections of the other detectors are suppressed by $nms$. The suppressed ones may be true detections. That is why top ranked false positives of the proposed method are mostly due to poor localization.

Fig.~\ref{fig:FPA} illustrates that the confusion with background error is significantly reduced. This shows the effectiveness of the proposed object likelihood features. Such strong object-saliency cues positively affect the proposed method to detect false detections.

Another observation shown in Fig.~\ref{fig:FPA} is that the proposed features could not reduce the confusion caused by similar object categories. However, they are effective on limiting the confusion between dissimilar object categories.

\subsection{Feature Importance}

\begin{table*}[t]
\begin{center}
\renewcommand\arraystretch{1.2}
\scalebox{0.84}{
\begin{tabular}{| l  l | c | c | c | c | c | c | c | c | c | c | c | c | c | c | c | c | c | c | c | c | c |}
\midrule[1.6pt]
\noalign{\vskip.2pt}
                     &  & aero & bike & bird & boa & bot & bus & car & cat & chr & cow & tab & dog & hor & mbik & pers & plnt & shp & sofa & tra & tv & mAP\\
\noalign{\vskip.2pt}
\midrule[1.6pt]
\noalign{\vskip.2pt}
\multirow{4}{*}{}

& $Rs$ & 	33.7 	 & 	62.4 	 & 	9.5 	 & 	15.2 	 & 	22.9 	 & 	50.4 	 & 	59.6 	 & 	18.8 	 & 	 22.0 	 & 	 22.8 	 & 	20.3 	 & 	6.2 	 & 	62.7 	 & 	53.7 	 & 	44.6 	 & 	 16.5 	 & 	24.4 	 & 	 34.4 	 & 	50.9 	 & 	 46.7 	 & 	33.9\\ \hline
& $Rs$ + $Os$  & 	35.6 	 & 	62.1 	 & 	10.6 	 & 	17.4 	 & 	24.6 	 & 	50.8 	 & 	59.3 	 & 	25.1 	 & 	 21.3 	 & 	23.2 	 & 	23.9 	 & 	10.5 	 & 	63.1 	 & 	51.0 	 & 	 45.5 	 & 	14.7 	 & 	26.3 	 & 	 37.8 	 & 	 50.6 	 & 	47.5 	 & 	35.1\\ \hline
& $R_S$ +  $So$   & 	35.4 	 & 	63.7 	 & 	10.6 	 & 	18.2 	 & 	26.5 	 & 	51.7 	 & 	60.3 	 & 	 18.7 	 & 	 22.7 	 & 	24.1 	 & 	21.5 	 & 	6.6 	 & 	63.8 	 & 	57.3 	 & 	 43.7 	 & 	18.5 	 & 	 24.3 	 & 	34.5 	 & 	 53.0 	 & 	45.3 	 & 	35.0\\ \hline
& All & 	36.8 	 & 	64.2 	 & 	12.3 	 & 	20.3 	 & 	27.3 	 & 	53.0 	 & 	60.3 	 & 	27.0 	 & 	 22.0 	 & 	 25.3 	 & 	27.1 	 & 	11.1 	 & 	63.7 	 & 	56.6 	 & 	45.4 	 & 	 19.3 	 & 	24.0 	 & 	 38.0 	 & 	54.5 	 & 	 46.8 	 & 	36.8\\

\noalign{\vskip.2pt}
\midrule[1pt]
\noalign{\vskip.2pt}

\end{tabular}
}
\end{center}

\caption{\label{table:FeatEff} The influence of selected features for the final detection performance.}
\end{table*}

\begin{figure}[t!]
\begin{center}
   \includegraphics[width=.75\linewidth]{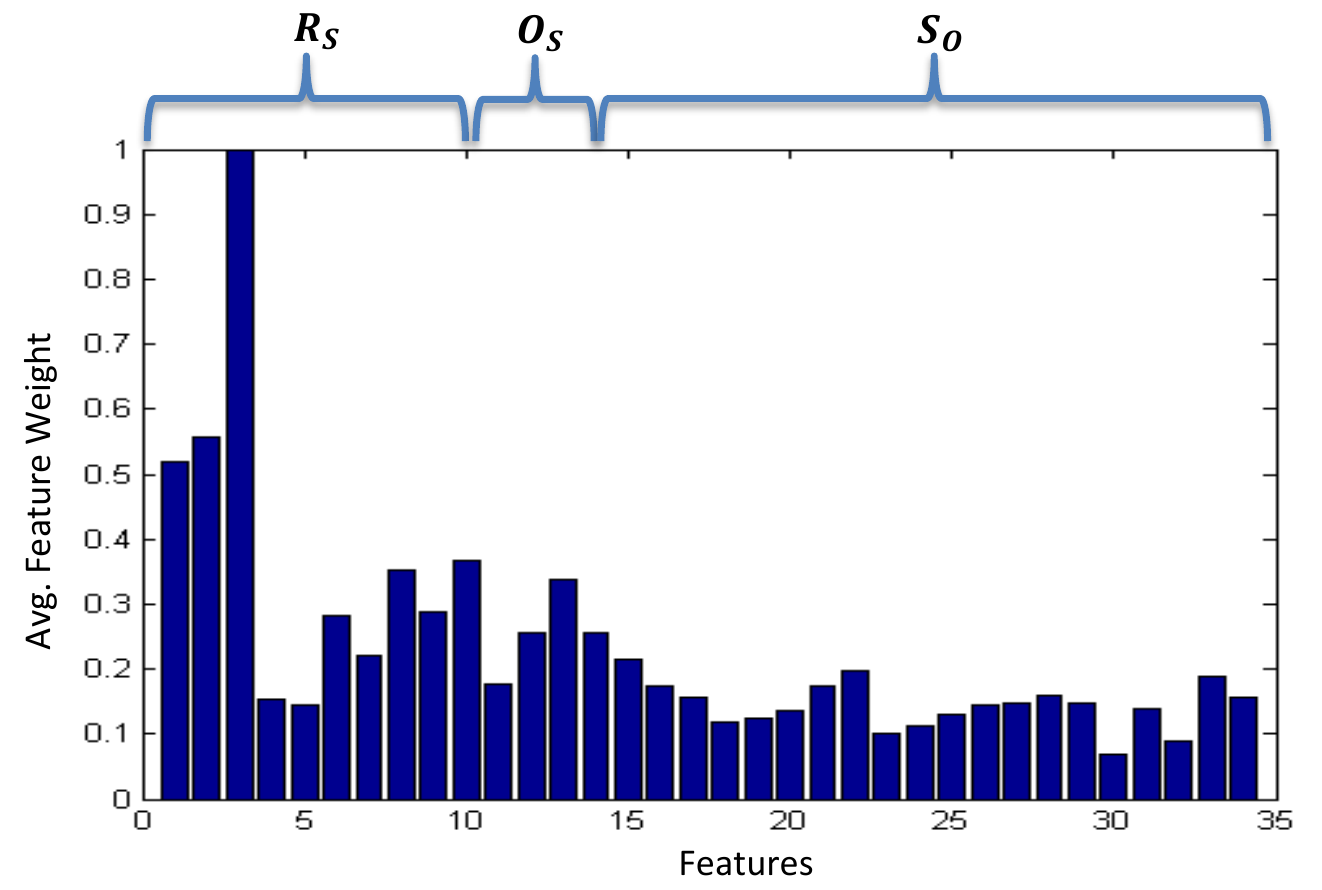}
\end{center}
   \caption{ Classifier weights are averaged over different classes to see the importance of features individually.}
\label{fig:featurewieghts}
\end{figure}

In this experiment, we study the influence of each individual feature. The weights are obtained by averaging the absolute classifier weights over the classes. The importance of proposed detector-detector context features ($R_S$ ) is highlighted in Fig.~\ref{fig:featurewieghts}. Moreover, feature weights also emphasize the importance of proposed object-saliency features ($O_S$). As stated earlier, the proposed $R_S$ and $O_S$ features are more generic and independent of the number of object categories. However, object-object relation exploited by other state-of-the-art context based object detection methods~\cite{hrContext,Desai,Cinbis} is dependent on the image characteristics. Therefore, the accuracy gain is limited to the image characteristics for these methods.

We now investigate the influence of each feature on the final mAP score. The detector scores are essential to rank the detection list. Therefore, it is not possible to evaluate $O_S$ and $S_O$ individually. We evaluate mAP using only $R_S$ feature. For the rest of the features, $R_S$ is also included. It is shown in Table~\ref{table:FeatEff} that using only $R_S$ improves the baseline detectors significantly. Object likelihood measure also improves the accuracy (e.g. for animal classes such as cat, dog or sheep). Significant improvement for these classes is due to poor representation capacity of template based detectors for non-rigid objects. Deformable part based object detectors are suited to detect rigid parts of the objects, see top ranked visual results of category cat in Fig.~\ref{fig:TFP}. Due to their (cats, dogs and sheeps) homogenous appearances, most of object proposals contain the full object shape. Therefore, detections of the entire objects receive higher confidence than detections for object parts. The object size plays role for the other animals, such as horse and cow. Object proposal methods used in this paper tend to perform better to cover small sized objects. Moreover, it is less likely to happen that object proposal methods generate many large bounding boxes for a specific image region. Therefore, the average overlap of a detection with these windows becomes lower. Adding object-object context ($S_O$) slightly improves most of the object classes. However, its contribution to the average precision increases when it is combined with the object-saliency. Furthermore, $S_O$ clearly improves the accuracy for class "bottle" in which samples usually occur within a context (usually on a table or in the hand of a person).

\begin{table*}[t]
\begin{center}
\renewcommand\arraystretch{1.2}
\scalebox{0.785}{
\begin{tabular}{| l  l | c | c | c | c | c | c | c | c | c | c | c | c | c | c | c | c | c | c | c | c | c | c |}
\midrule[1.6pt]
\noalign{\vskip.2pt}
                     &  & aero & bike & bird & boa & bot & bus & car & cat & chr & cow & tab & dog & hor & mbik & pers & plnt & shp & sofa & tra & tv & mAP & Imp.\\
\noalign{\vskip.2pt}
\midrule[1.6pt]
\noalign{\vskip.2pt}
\multirow{3}{*}{}

&DPM + Context                  & 	29.8 	 & 	57.5   & 	9.9 	 & 	16.4 	 & 	24.1 	 & 	46.3 	 & 	 58.0 	 & 	 21.1 	 & 	19.6 	 & 	20.4 	 & 	15.1 	 & 	7.5 	 & 	58.3 	 & 	 50.4 	 & 	42.0 	 & 	 14.3 	 & 	18.2 	 & 	 28.0 	 & 	49.0 	 & 	39.6 &	31.3 &  3.0 \\ \hline
&CN + Context  & 	33.3 	 & 	55.1  & 	11.4 	 & 	13.4 	 & 	22.7 	 & 	44.9 	 & 	57.0 	 & 	22.6 	 & 	 18.6 	 & 	19.4 	 & 	17.5 	 & 	8.5 	 & 	56.0 	 & 	50.9 	 & 	 42.1 	 & 	17.4 	 & 	20.9 	 & 	 31.3 	 & 	 48.5 	 & 	45.5 &  31.9 & 1.9 \\ \hline
&EES + Context                 & 	31.4 	 & 	57.2   & 	10.6 	 & 	16.9 	 & 	21.0 	 & 	46.6 	 & 	 51.5 	 & 	 13.3 	 & 	15.5 	 & 	20.6 	 & 	15.2 	 & 	8.1 	 & 	57.3 	 & 	 51.5 	 & 	32.9 	 & 	 14.1 	 & 	18.0 	 & 	 20.1 	 & 	46.9 	 & 	44.5 	 & 29.7 & 10.0\\  \hline
\noalign{\vskip.2pt}

\end{tabular}
}
\end{center}
\caption{\label{table:Single} The results of the re-ranked SINGLE baseline detector outputs using contextual features. The results of SINGLE detectors are improved using context.}
\end{table*}

\begin{table*}[t]
\begin{center}
\renewcommand\arraystretch{1.2}
\scalebox{0.845}{
\begin{tabular}{| l  c | c | c | c | c | c | c | c | c | c | c | c | c | c | c | c | c | c | c | c | c | c |}
\midrule[1.6pt]
\noalign{\vskip.2pt}
                     &  & aero & bike & bird & boa & bot & bus & car & cat & chr & cow & tab & dog & hor & mbik & pers & plnt & shp & sofa & tra & tv & mAP\\
\noalign{\vskip.2pt}
\midrule[1.6pt]
\noalign{\vskip.2pt}
\multirow{9}{*}{}

&BL1             & 28.6 & 55.1 & 0.6  & 14.5 & 26.5 & 39.7 & 50.1 & 16.5 & 16.5 & 16.8 & 24.6 & 5.0  & 45.2 & 38.3 & 35.8 & 9.0  & 17.4 & 22.7 & 34.0 & 38.3 & 26.8\\ \hline
&\cite{Desai}    & 1.7  & 0    & 0.1  & 1.4  &  0   & -3.5 & 1.3  & 0.5  & -2.8 & 1.2  & -0.7 & 0.2  & 0.5  & 1.1  & -2.8 & -1.1 & -2.3 & -0.7 & 0.5  &   0  & -0.3\\ \hline
&\cite{hrContext}& 2.4  & -4.2 & 2.3  & 0.8  & -1.1 & -0.2 & -0.4 & 3.9  & 1.6  & 0.9  & 2.3  & 6.9  & 5.6  & 2.2  &  0   & 4.7  & 3.8  &  2.8 & 4.7  & -0.1 & 1.9 \\ \hline
&\cite{Cinbis}   & 5.6  & 2.7  & 9.2  & 0.8  & 3.2  & 1.9  & 3.4  & 5.0  &  0   & 0.7  & 1.4  & 7.9  & 5.9  & 4.6  & 3.5  & 4.2  & 3.1  & 4.9  & 4.9  & 0.3  & 3.6\\ \hline
\midrule[1.6pt]
&BL2             & 27.8 & 55.9 & 1.4  & 14.6 & 25.7 & 38.1 & 47.0 & 15.1 & 16.3 & 16.7 & 22.8 & 11.1  & 43.8 & 37.3 & 35.2 & 14.0  & 16.9 & 19.3 & 31.9 & 37.3 & 26.4\\ \hline
&\cite{Yukun1}    & 2.4 & 1.9 & 0.5  & 0.2 & 3.2 & 2.6 & 2.9 & -0.9 & 0.9 & 1.9 & 0.2 & 5.3 & 1.3 & 3.3 & 3.6 & 3.0 & 3.2 & 3.7 & 2.9 & -0.5 & 2.0\\ \hline
\midrule[1.6pt]
&BL3             & 26.7 & 56.9 & 2.6  & 12.8 & 	21.9  & 46.0 & 	55.3 & 	13.7  & 19.0 & 	19.4 & 	12.6 & 2.2 & 58.1 & 47.3 & 40.9 & 6.8 & 15.0 & 	26.9 & 	43.4 & 	38.8 & 	28.3\\ \hline
&Proposed             & 	10.1 	 & 	7.3 	 & 	9.7 	 & 	7.5 	 & 	5.4 	 & 	7.0 	 & 	5.0 	 & 	 13.3 	 & 	3.0 	 & 	5.9 	 & 	14.5 	 & 	8.9 	 & 	5.6 	 & 	9.3 	 & 	4.5 	 & 	12.5 	 & 	8.9 	 & 	11.0 	 & 	11.1 	 & 	8.1 	 &  8.4\\ \hline
\noalign{\vskip.2pt}

\end{tabular}
}
\end{center}
\caption{\label{table:Context} Comparison of the state-of-the art context based object detection methods on PASCAL VOC07 dataset. The results of referred works~\cite{Desai,hrContext,Cinbis} and $DPM$ baseline scores ($BL1$) are reported in~\cite{Cinbis} whereas ~\cite{Yukun1} and $DPM$ baseline scores ($BL2$) are reported in~\cite{Yukun1}. $BL3$ is $DPM$ baseline score obtained in this paper. The results represented as proposed are the improvements over $DPM$ baseline in this paper.}
\end{table*}

\begin{table*}[t]
\begin{center}
\renewcommand\arraystretch{1.2}
\scalebox{0.81}{
\begin{tabular}{| l  c | c | c | c | c | c | c | c | c | c | c | c | c | c | c | c | c | c | c | c | c | c |}
\midrule[1.6pt]
\noalign{\vskip.2pt}
                     &  & aero & bike & bird & boa & bot & bus & car & cat & chr & cow & tab & dog & hor & mbik & pers & plnt & shp & sofa & tra & tv & mAP\\
\noalign{\vskip.2pt}
\midrule[1.6pt]
\noalign{\vskip.2pt}
\multirow{9}{*}{}

&BL1 & 	46.3 	 & 	49.5	 & 	4.8 	 & 	6.4 	 & 	22.6 	 & 	53.5 	 & 	38.7 	 & 	24.8 	 & 	14.2 	 & 	 10.5 	 & 	10.9 	 & 	12.9 	 & 	36.4 	 & 	38.7 	 & 	 42.6 	 & 	3.6 	 & 	26.9 	 & 	22.7 	 & 	 34.2 	 & 	 31.2 	 & 	26.6\\ \hline
& DPM-Context\cite{DPM} & 	0.1 	 & 	1.3	 & 	2.7 	 & 	 1.8	 & 	-0.6 	 & 	1.8	 & 	2.9  & 	-4.8 & 	0.5  & 	 1.3 & 	0.7 & 	1.0 & 	1.5 & 	1.5	 & 	2.5 	 & 0.6	 & 	-2.8 & 	4.9  & 	 6.6 	 & 	2.7 & 	1.2\\ \hline
&\cite{InWild} & 	6.5 	 & 	-0.7	 & 	7.2 	 & 	4.4 	 & 	6.5 	 & 	1.7 	 & 	6.9 	 & 	7.2 	 & 	 0.0 	 & 	2.1 	 & 	2.8 	 & 	3.7 	 & 	3.4 	 & 	5.5 	 & 	 2.5 	 & 4.6	 & 	8.4 & 	3.3  & 	 7.9 	 & 	 3.1 	 & 	4.2\\ \hline
\midrule[1.6pt]
&BL2         &  37.4 	 & 	51.8 	 & 	5.1 	 & 	3.9 	 & 	20.3 	 & 	51.4 	 & 	39.2 	 & 	13.3 	 & 	 15.2 	 & 	9.5 	 & 	7.2 	 & 	4.8 	 & 	40.1 	 & 	 43.4 	 & 	 41.5 	 & 	9.8 	 & 	13.2 	 & 	 16.4 	 & 	31.9 	 & 	26.5 	 & 	24.1\\ \hline
&Proposed    & 	7.3 	 & 	2.9 	 & 	8.5 	 & 	6.6 	 & 	2.5 	 & 	7.1 	 & 	5.1 	 & 	15.0 	 & 	 2.8 	 & 	3.5 	 & 	5.8 	 & 	7.1 	 & 	3.6 	 & 	7.7 	 & 	 3.5 	 & 	8.0 	 & 	8.4 	 &  3.4 	 & 	10.6 	 & 	11.8 	 & 	6.6 \\ \hline
\noalign{\vskip.2pt}

\end{tabular}
}
\end{center}
\caption{\label{table:Context10} Comparison of the state-of-the art context based object detection methods on PASCAL VOC10$val$. The results of referred works~\cite{InWild} and $DPM$ baseline scores ($BL1$) are reported in~\cite{InWild}. $BL2$ is $DPM$ baseline score obtained in this paper. The results represented as proposed are the improvements over $DPM$ baseline in this paper.}
\end{table*}

\subsection{Re-ranking Detections from Single Detector}
\begin{table*}[t]
\begin{center}
\renewcommand\arraystretch{1.2}
\scalebox{0.86}{
\begin{tabular}{| l  c | c | c | c | c | c | c | c | c | c | c | c | c | c | c | c | c | c | c | c | c | c |}
\midrule[1.6pt]
\noalign{\vskip.2pt}
                     &  & aero & bike & bird & boa & bot & bus & car & cat & chr & cow & tab & dog & hor & mbik & pers & plnt & shp & sofa & tra & tv & mAP\\
\noalign{\vskip.2pt}
\midrule[1.6pt]
\noalign{\vskip.2pt}
\multirow{11}{*}{}

&~\cite{DPM}  & 	36.8 	 & 	50.1 	 & 	4.3 	 & 	10.6 	 & 	14.3 	 & 	50.0 	 & 	40.4 	 & 	13.9 	 & 	 15.9 	 & 	14.2 	 & 	9.4 	 & 	4.7 	 & 	41.8 	 & 	43.0 	 & 	 40.9 	 & 	5.9 	 & 	11.6 	 & 	 15.3 	 & 	 33.4 	 & 	31.4 	 & 	24.4\\ \hline
&~\cite{CNHOG} & 	34.5 	 & 	48.8 	 & 	5.3 	 & 	10.4 	 & 	11.4 	 & 	52.1 	 & 	40.9 	 & 	18.7 	 & 	 14.9 	 & 	15.7 	 & 	7.1 	 & 	5.9 	 & 	41.3 	 & 	45.5 	 & 	 42.2 	 & 	10.1 	 & 	14.0 	 & 	 18.1 	 & 	 36.2 	 & 	35.8 	 & 	25.4 \\ \hline
&~\cite{EES}  & 	22.6 	 & 	34.9 	 & 	3.2 	 & 	9.4 	 & 	4.5 	 & 	45.9 	 & 	25.0 	 & 	2.1 	 & 	 7.2 	 & 	10.7 	 & 	4.3 	 & 	2.0 	 & 	21.7 	 & 	31.7 	 & 	 10.0 	 & 	2.1 	 & 	11.6 	 & 	 8.1 	 & 	 21.3 	 & 	23.6 	 & 	15.1\\ \hline
& PoW2        & 	44.8 	 & 	53.3 	 & 	14.3 	 & 	14.6 	 & 	14.2 	 & 	56.3 	 & 	44.7 	 & 	27.2 	 & 	 18.9 	 & 	19.6 	 & 	14.5 	 & 	15.0 	 & 	44.1 	 & 	50.0 	 & 	 45.4 	 & 	13.2 	 & 	17.6 	 & 	 22.5 	 & 	 42.0 	 & 	39.1 	 & 	30.6\\ \hline
\midrule[1.6pt]
&~\cite{DPM}  & 	37.4 	 & 	51.8 	 & 	5.1 	 & 	3.9 	 & 	20.3 	 & 	51.4 	 & 	39.2 	 & 	13.3 	 & 	 15.2 	 & 	9.5 	 & 	7.2 	 & 	4.8 	 & 	40.1 	 & 	43.4 	 & 	 41.5 	 & 	9.8 	 & 	13.2 	 & 	 16.4 	 & 	 31.9 	 & 	26.5 	 & 	24.1\\ \hline
&~\cite{CNHOG} & 	36.6 	 & 	45.0 	 & 	6.0 	 & 	4.7 	 & 	17.9 	 & 	52.5 	 & 	40.2 	 & 	18.8 	 & 	 15.3 	 & 	10.6 	 & 	6.5 	 & 	5.2 	 & 	39.7 	 & 	44.4 	 & 	 44.0 	 & 	15.5 	 & 	16.4 	 & 	 13.0 	 & 	 35.6 	 & 	33.8 	 & 	25.1\\ \hline
&~\cite{EES}  & 	19.9 	 & 	36.8 	 & 	1.8 	 & 	3.3 	 & 	7.2 	 & 	46.2 	 & 	23.5 	 & 	2.0 	 & 	 4.2 	 & 	6.4 	 & 	2.1 	 & 	1.3 	 & 	20.6 	 & 	30.4 	 & 	9.5 	 & 	2.8 	 & 	14.5 	 & 	 7.0 	 & 	 24.0 	 & 	24.7 	 & 	14.4 \\ \hline
& PoW2        & 	44.7 	 & 	54.7 	 & 	13.6 	 & 	10.5 	 & 	22.8 	 & 	58.5 	 & 	44.3 	 & 	28.3 	 & 	 18.0 	 & 	12.9 	 & 	13.0 	 & 	11.9 	 & 	43.7 	 & 	51.0 	 & 	 45.0 	 & 	17.8 	 & 	21.6 	 & 	 19.8 	 & 	 42.5 	 & 	38.2 	 & 	30.7 \\ \hline
\noalign{\vskip.2pt}

\end{tabular}
}
\end{center}
\caption{\label{table:Result10} The results for baselines ($DPM$, $CN$ and $EES$) and proposed detector merging scheme using PoW2 on VOC10 (upper: $train$ set and lower: $val$ set).}
\end{table*}

In this experiment, we exploit the effectiveness of context features without combining detectors into a single list. The proposed context features are only used to re-rank individual detectors. It is shown in Table~\ref{table:Single} that the proposed method is still effective and improves the baseline detectors. However, the accuracy gain is relatively smaller than using the combined detector outputs in Table~\ref{table:Result}. These results underline the importance of combining different detector outputs to recover from missed detections to improve the overall object detection performance.

Note that a detector with a high recall and low precision such as $EES$ can be as powerful as other more precise detectors ($DPM$, $CN$) using the proposed context features.

\subsection{Comparison to Other Context Methods:}
In this experiment, we compare the proposed method against the state-of-the-art context based object detection re-ranking methods. Table~\ref{table:Context} shows the baseline scores of DPM and improvements reported by the papers~\cite{Cinbis,Yukun1} on VOC07. The gain in performance by our method indicates the importance of high level contextual features and L2R based detector merging.

Moreover, the proposed method is compared to the recent work by Mottaghi et al.~\cite{InWild} on VOC10 dataset (See Table~\ref{table:Context10}). The authors report also context re-ranking method of $DPM$ (See~\cite{DPM} for details) discussed in Section~\ref{relatedWork}.

The contextual features proposed by other methods in Table~\ref{table:Context} and Table~\ref{table:Context10} are from different sources. Hence, they can be complementary to the proposed features. Combination of these features may further improve the results.

\subsection{Tests on VOC10}
We also evaluate our method on the PASCAL VOC10 dataset. The VOC10 annotations of the test samples are not publicly available. Therefore, we use only the "$train/val$" dataset. All the training is done on the VOC07 $trainval$ set, including object detection models and detector-detector relation models. Table~\ref{table:Result10} shows the results. Table~\ref{table:Result10} indicates that the proposed method outperforms the baseline detectors for all classes also on the cross dataset evaluation. The results show that the learned detector-detector context is generic and it is not dataset dependent.

\begin{figure}[h!]
\begin{center}
   \includegraphics[width=1\linewidth]{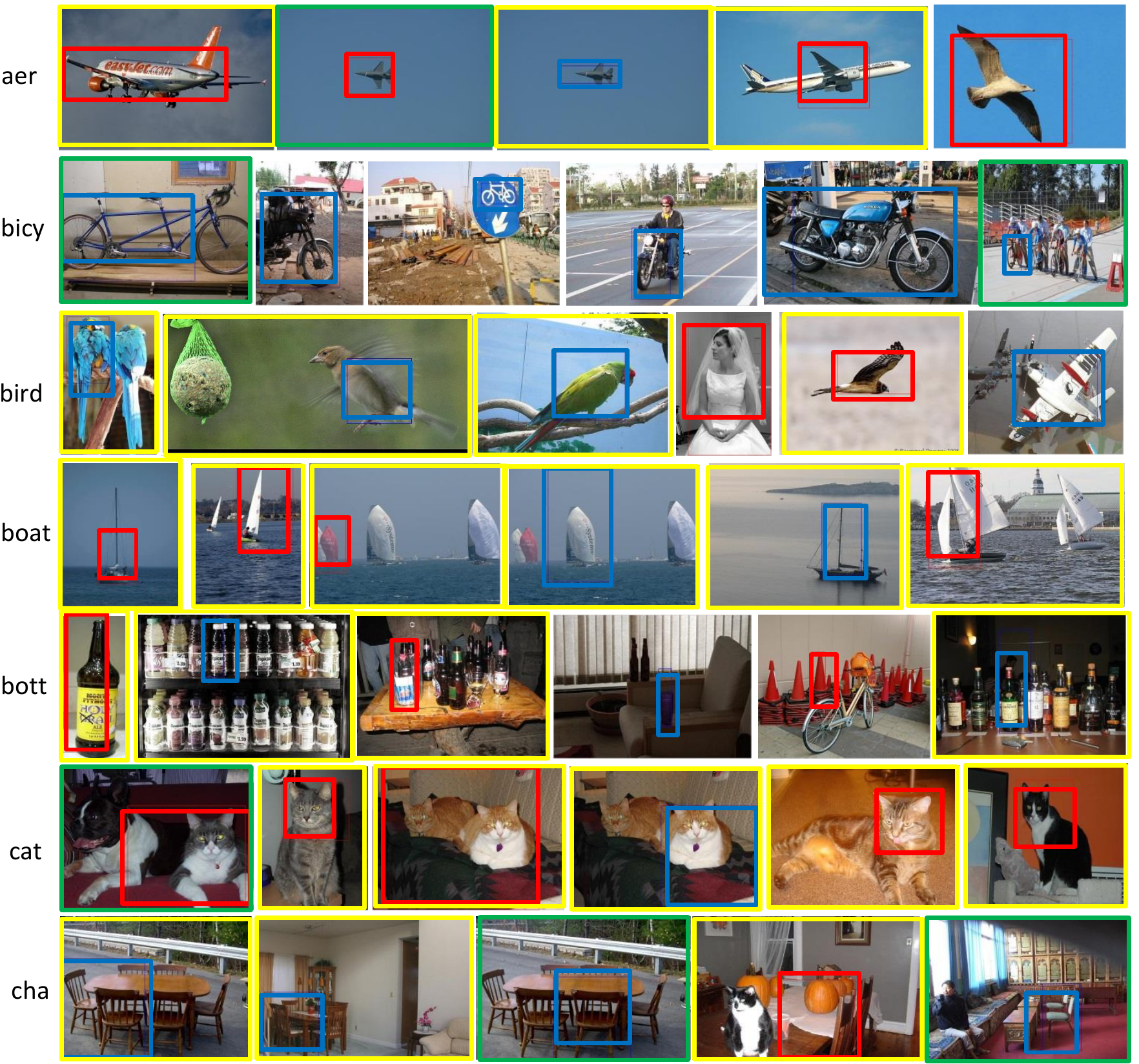}
\end{center}
   \caption{Top ranked false positives of the proposed method for specified classes. Blue and red colors indicate the detector type, $DPM$ and $CN$, respectively. Yellow and green colors correspond to poor localization and multiple detections, respectively. The image frames without color information indicates no overlap between ground truth object, either due to miss classification or background clutter.}
   \label{fig:TFP}
\end{figure}

\begin{figure}[h]
\begin{center}
   \includegraphics[width=1\linewidth]{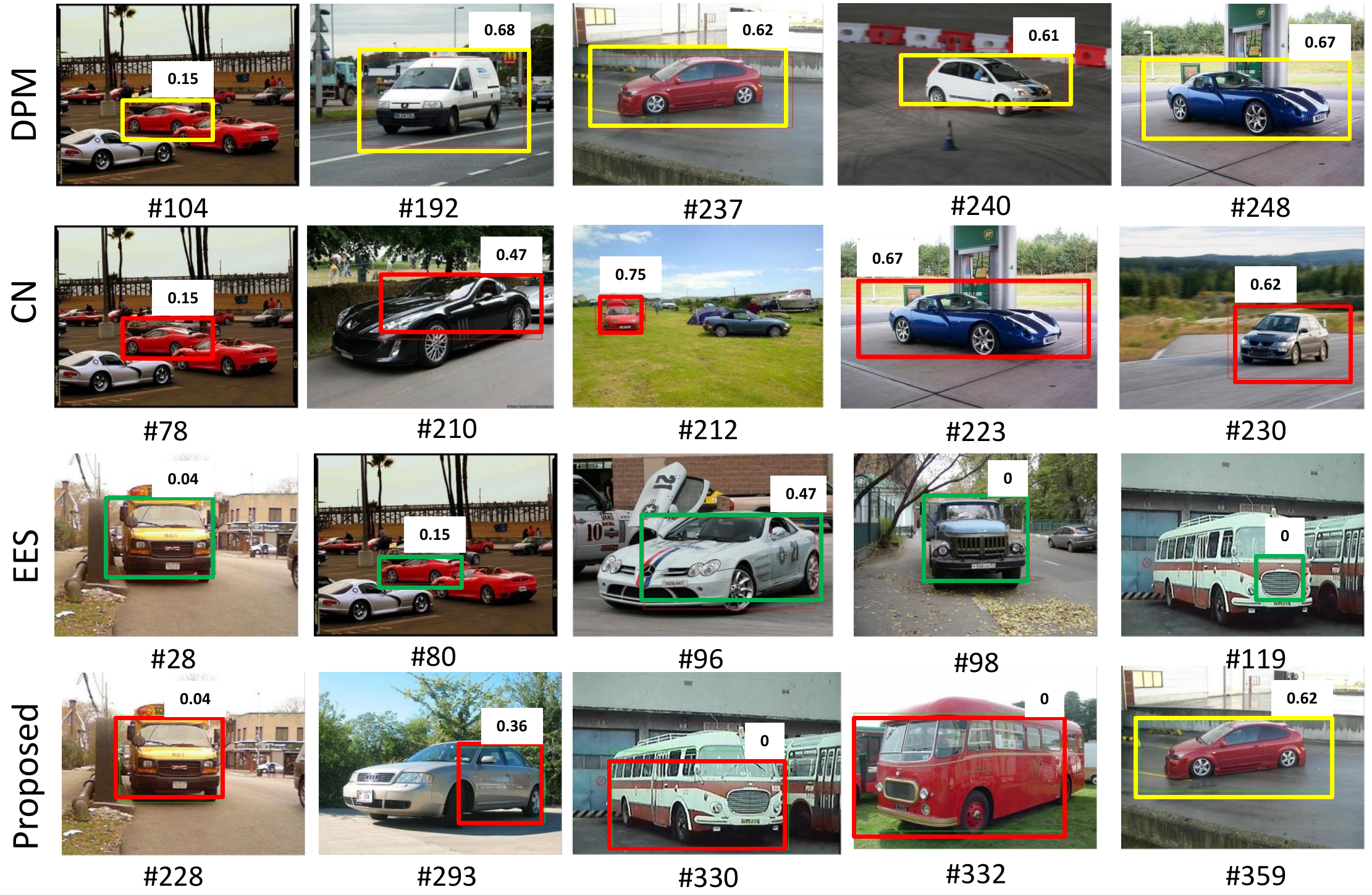}
\end{center}
   \caption{Top five ranked false positives of baseline detectors and proposed method with their rankings below (object class car). The color of the detection yellow, green and red indicates the type of detector $DPM$, $CN$ and $EES$ respectively. The false positives of individual detectors are pushed down in the proposed method.}
\label{fig:discuss10}
\end{figure}

\section{Discussion}

Diversity, and thus potential complementary detections of detectors exist mainly due to two reasons. The first one is related to the features used to represent the model. Although the selected detectors in this paper use HOG based features. They are differentiated by additional color features or feature extraction steps performed at different scales. The second source of the diversity comes from the classifier. These differences have a substantial influence on their final outputs. Table~\ref{table:Bound} and Table~\ref{table:Result} represent the differences in the final outputs. Although these detectors have detections in common, they are complementary to each other. While common detections are useful to learn consistency in their output, complementary detections resolves missed detections for each individual detector. A higher diversity in detectors will further improve the results.

With the help of the proposed method, future object detectors can focus on more specific solutions to harder detection problems. Their results will be combined with other detection methods to carry object detection algorithms a step further. The contribution of new method can be measured compared to the combination of the-state-of-the-art methods.

The pointwise and pairwise approaches do not consider specific properties of ranking. For instance, position is not used in the loss function. Enabling the position information in the loss function may further improve the final results. However, the trade-off between the complex structure and the accuracy should be taken into account.

To avoid overfitting, the object detectors are trained on $train$ to test on $val$. Further, they are trained on $val$ to test on $train$. The detectors are trained with fewer examples. This has an impact on the performance of detectors on $train/val$ set in which we learn the relationship between detectors. It is observed that for some classes the performance of object detectors on $train/val$ set are not inline with $test$ set. Therefore, learning the models for detectors on a larger dataset may further improve the proposed learning to rank scheme.

The non-maximum suppression technique is a widely used ad-hoc method in object detection literature. However, learning to detect multiple detections from different detectors may be more appropriate for the proposed method.

The proposed method does not provide new bounding boxes. Therefore, it cannot recover from poor localization errors. Poor localization error becomes problematic for some cases (See Fig.~\ref{fig:TFP} and Fig.~\ref{fig:discuss10} for top ranked false positives). This problem can be resolved by proposing new bounding boxes using object proposals or using a method similar to~\cite{RCNN}.

\section{Conclusion}

No detection algorithm can be considered as universal. As a consequence, we have proposed an approach to combine different object detectors. The proposed approach uses (single) object detectors to exploit their correlation by learning a re-ranking scheme.

The proposed method uses common detections of single detectors to award a detection based on detector correlation and consistency. Whereas the proposed method uses complementary detections of detectors to recover missed detections of each single detector.

Experiments on the PASCAL VOC07 and VOC10 datasets show that the proposed method significantly outperforms individual object detectors, $DPM$ (8.4\%), $CN$ (6.8\%) and $EES$ (17.0\%) on VOC07 and $DPM$ (6.5\%), $CN$ (5.5\%) and $EES$ (16.2\%) on VOC10.


\begin{figure*}[h]
\begin{center}
   \includegraphics[width=1\linewidth]{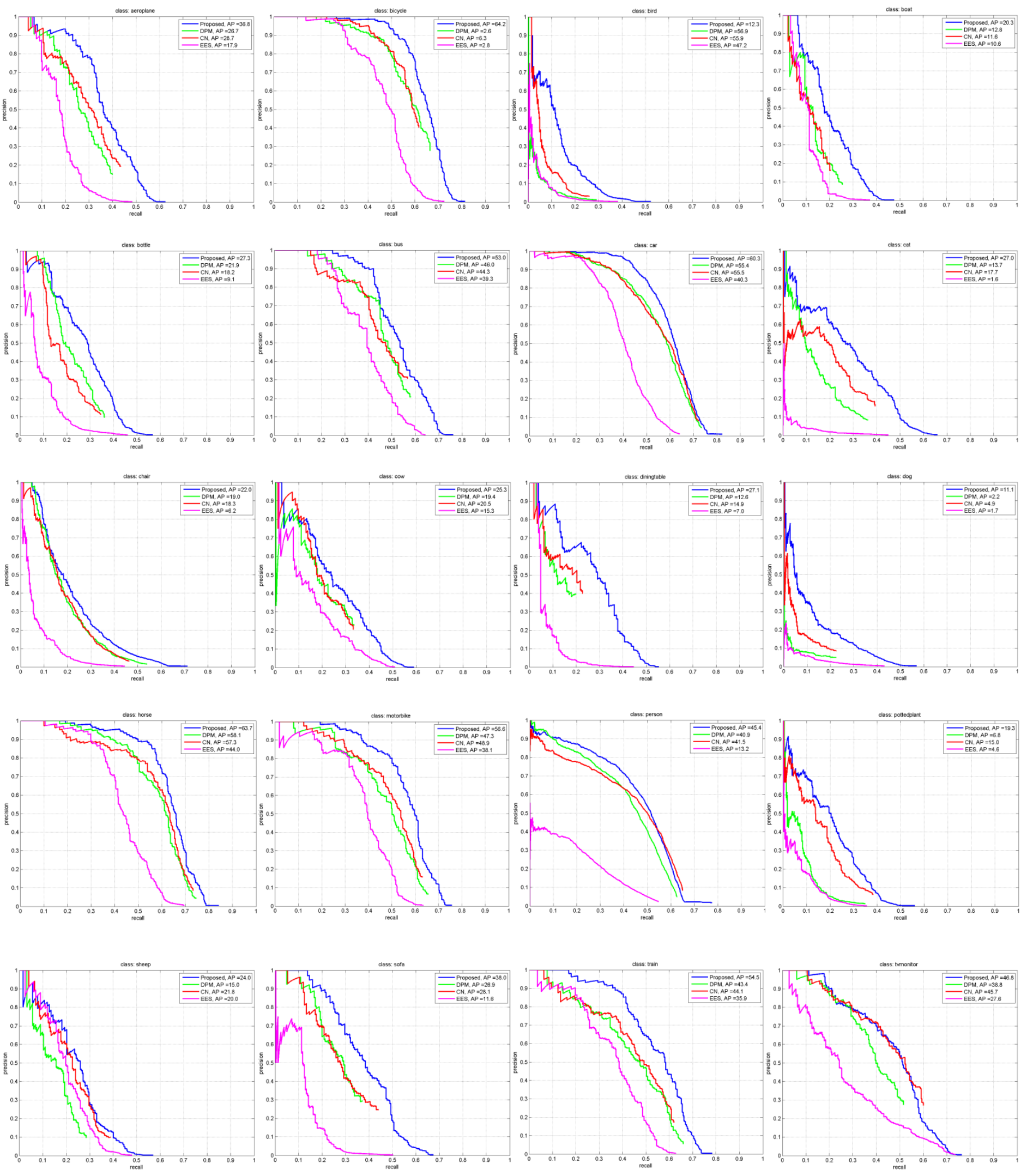}
\end{center}
   \caption{Precision-recall curves on PASCAL VOC 2007. The proposed method significantly outperforms all single detectors. Furthermore, it is shown that detections of baseline detectors have remarkable differences.}
\label{fig:flow}
\label{fig:discuss2}
\end{figure*}

%

\appendices




\ifCLASSOPTIONcaptionsoff
  \newpage
\fi



\end{document}